\documentclass[11pt,a4paper]{article}
\usepackage[top=3cm, bottom=3cm, left=3cm, right=3cm]{geometry}

\usepackage{booktabs}

\usepackage{color}
\usepackage{latexsym}              
\usepackage{amsmath}               
\usepackage{amssymb}               
\usepackage{amsfonts}              
\usepackage{amsthm}                
\usepackage{multirow}
\usepackage{tikz}                  
\usetikzlibrary{arrows,positioning,shapes}
\usepackage{tcolorbox}
\usepackage{times}
\usepackage[utf8]{inputenc}
\usepackage[T1]{fontenc}
\usepackage{enumitem}
\usepackage{xcolor}
\usepackage{thmtools}
\usepackage{thm-restate}
\usepackage[square,numbers]{natbib}
\usepackage[colorlinks=true,allcolors=blue]{hyperref}
\usepackage{cleveref}

\renewenvironment{proof}[1][\proofname]{{\bfseries #1.}}{}

\makeatother

 \newtheorem{example}{Example}
  \newtheorem{theorem}{Theorem}
  \newtheorem{lemma}{Lemma}
  \newtheorem{proposition}{Proposition}
  \newtheorem{remark}{Remark}
  \newtheorem{corollary}{Corollary}
  \newtheorem{definition}{Definition}

\newcommand{\eqals}[1]{\begin{align*}#1\end{align*}}
\newcommand{\eqal}[1]{\begin{align}#1\end{align}}
\newcommand{\bpr}{\begin{proof}}
\newcommand{\epr}{\end{proof}}
\newcommand{\be}{\begin{equation}}
\newcommand{\ee}{\end{equation}}

\newcommand{\bd}{\begin{definition}}
\newcommand{\ed}{\end{definition}}

\newcommand{\bi}{\begin{itemize}}
\newcommand{\ei}{\end{itemize}}

\newtheorem{ass}{Assumption}
\newcommand{\ba}{\begin{ass}}
\newcommand{\ea}{\end{ass}}

\newcommand{\bre}{\begin{restatable}}
\newcommand{\ere}{\end{restatable}}

\newcommand{\br}{\begin{remark}}
\newcommand{\er}{\end{remark}}

\newcommand{\bp}{\begin{proposition}}
\newcommand{\ep}{\end{proposition}}

\newcommand{\blm}{\begin{lemma}}
\newcommand{\elm}{\end{lemma}}

\newcommand{\bt}{\begin{theorem}}
\newcommand{\et}{\end{theorem}}

\newcommand{\bcor}{\begin{corollary}}
\newcommand{\ecor}{\end{corollary}}

\newcommand{\bex}{\begin{example}}
\newcommand{\eex}{\end{example}}

\crefname{proposition}{Proposition}{Propositions}

\crefname{definition}{Definition}{Definitions}

\crefname{ass}{Assumption}{Assumptions}
\crefname{equation}{Eq.}{Eqs.}
\crefname{figure}{Fig.}{Figs.}
\crefname{table}{Table}{Tables}
\crefname{section}{Sec.}{Secs.}

\crefname{theorem}{Thm.}{Thms.}

\crefname{lemma}{Lemma}{Lemmas}

\crefname{corollary}{Cor.}{Cors.}

\crefname{example}{Example}{Examples}

\crefname{appendix}{Appendix}{Appendixes}

\crefname{remark}{Remark}{Remark}

\newcommand{\Deff}{\mathsf{df}}

\newcommand{\Exp}[1]{\mathbb{E}\left[#1 \right]}

\newcommand{\Expb}[2]{\mathbb{E}_{#1}\left[#2\right]}
\newcommand{\Hess}{\mathbf{H}}

\newcommand{\biass}{{\mathsf{Bias}}}
\newcommand{\clas}{\mathcal{P}}

\newcommand{\Bone}{\mathsf{B}_1}
\newcommand{\Btwo}{\mathsf{B}_2}

\newcommand{\Bos}{\mathsf{B}_1^\star}
\newcommand{\Bts}{\mathsf{B}_2^\star}

\newcommand{\Qs}{\mathsf{Q}^\star}
\newcommand{\Lc}{\mathsf{L}}
\newcommand{\Fc}{\mathsf{F}}
\newcommand{\Qc}{\mathsf{Q}}
\newcommand{\Cb}{\mathsf{C}_{\textup{bias}}}
\newcommand{\Cv}{\mathsf{C}_{\textup{var}}}
\newcommand{\Cla}{\mathsf{C}_\la}

\newcommand{\Kfinal}{\mathsf{K}}
\newcommand{\Kv}{K_{\textup{var}}}
\newcommand{\Deffsup}{\overline{\Deff}}

\newcommand{\Bob}{\overline{\mathsf{B}}_1}
\newcommand{\Btb}{\overline{\mathsf{B}}_2}
\newcommand{\rhohat}{\widehat{\rho}}
\newcommand{\rla}{\mathsf{r}_\la}
\newcommand{\tla}{\mathsf{t}_\la}
\newcommand{\tlab}{\widetilde{\mathsf{t}}_{\lambda}}
\newcommand{\nm}[1]{\mathsf{t}(#1)}
\newcommand{\nmm}[1]{\mathsf{t}^{\mu}(#1)}
\newcommand{\Rad}{R}

\newcommand{\X}{\mathcal{X}}
\newcommand{\Y}{\mathcal{Y}}
\newcommand{\Z}{\mathcal{Z}}
\newcommand{\R}{\mathbb{R}}

\newcommand{\N}{\mathbb{N}}

\newcommand{\E}{\mathbb{E}}
\newcommand{\Lo}{L}
\renewcommand{\L}{L}
\newcommand{\Ll}{L_{\lambda}}

\newcommand{\Lnl}{\widehat{L}_{\lambda}}

\newcommand{\Ho}{\mathbf{H}}
\newcommand{\Hl}{\mathbf{H_{\lambda}}}

\newcommand{\Hnl}{\mathbf{\widehat{H}_{\lambda}}}

\newcommand{\othet}{\theta^{\star}}
\newcommand{\othetl}{\theta^\star_{\lambda}}
\newcommand{\othetnl}{\widehat{\theta}^\star_{\lambda}}

\newcommand{\Finf}{\mathcal{F}_{\infty}}

\newcommand{\la}{\lambda}
\newcommand{\hh}{{\mathcal{H}}}

\DeclareMathOperator{\supp}{supp}

\DeclareMathOperator{\Tr}{Tr}

\DeclareMathOperator{\poly}{poly}

\newcommand{\dikin}{\underline{\phi}}

\newcommand{\dikins}{\overline{\phi}}

\newcommand{\bvar}{\widehat{\mathsf{Var}}}

\DeclareMathOperator*{\argmin}{arg\,min}

\renewcommand{\leq}{\leqslant}
\renewcommand{\geq}{\geqslant}

\date{}

\begin{document}

\title{Beyond Least-Squares: Fast Rates for Regularized Empirical Risk Minimization through Self-Concordance}

\author{\\
Ulysse Marteau-Ferey ~~~~Dmitrii Ostrovskii ~~~~Francis Bach ~~~~Alessandro Rudi \\ [3ex]
INRIA - D\'epartement d'Informatique de l'\'Ecole Normale Sup\'erieure \\
PSL Research University \\
Paris, France
}


\maketitle

\begin{abstract}%
We consider learning methods based on the regularization of a convex empirical risk by a squared Hilbertian norm, a setting that includes linear predictors and non-linear predictors through positive-definite kernels. In order to go beyond the generic analysis leading to convergence rates of the excess risk as $O(1/\sqrt{n})$ from $n$ observations, we assume that the individual losses are self-concordant, that is, their third-order derivatives are bounded by their second-order derivatives. This setting includes least-squares, as well as all generalized linear models such as logistic and softmax regression. For this class of losses, we provide a bias-variance decomposition and show that the assumptions commonly made in least-squares regression, such as the source and capacity conditions, can be adapted to obtain fast non-asymptotic rates of convergence by improving the bias terms, the variance terms or both.
\end{abstract}

\section{Introduction}

Regularized empirical risk minimization remains a cornerstone of statistics and supervised learning, from the early days of linear regression~\citep{hoerl1976ridge} and neural networks~\citep{geman1992neural}, then to spline smoothing~\citep{wahba1990spline} and more generally kernel-based methods~\citep{shawe2004kernel}. While the regularization by the squared Euclidean norm is applied very widely, the statistical analysis of the resulting learning methods is still not complete.

The main goal of this paper is to provide a sharp non-asymptotic analysis of regularized empirical risk minimization (ERM), or more generally regularized $M$-estimation, that is estimators obtained as the unique solution of 
    \vspace{-4pt}  \begin{equation}
  \displaystyle \min_{\theta \in \mathcal{H}}\ \  \frac{1}{n} \sum_{i=1}^n \ell_{z_i}(\theta) + \frac{\la}{2}\| \theta\|^2,
   \vspace{-4pt} 
   \end{equation}
    where $\mathcal{H}$ is a Hilbert space (possibily infinite-dimensional) and $\ell_z(\theta)$ is the convex loss associated with an observation $z$ and the estimator $\theta \in \mathcal{H}$. We assume that the observations $z_i$, $i=1,\dots,n$ are independent and identically distributed, and that the minimum of the associated unregularized expected risk $L(\theta)$ is attained at a certain $\othet \in \mathcal{H}$.
    
In this paper, we focus  on dimension-independent results \citep[thus ultimately extending the  analysis in the finite-dimensional setting from][]{ostrovskii2018finite}. For this class of problems, two main classes of problems have been studied, depending on the regularity assumptions on the loss.

Convex Lipschitz-continuous losses (with respect to the parameter $\theta$), such as for logistic regression or the support vector machine, lead to general \emph{non-asymptotic} bounds for the excess risk of the form~\citep{FR}:
    \vspace{-1pt}
\begin{equation}
\label{FR}
\frac{B^2}{\la n} + \la \| \othet\|^2,
    \vspace{-1pt}
\end{equation}
where $B$ is a uniform upper bound on the Lipschitz constant for all losses $\theta \mapsto \ell_z(\theta)$. The bound above already has a form that takes into account two separate terms: a \emph{variance term} ${B^2}/({\la n})$ which depends on the sample size $n$ but not on the optimal predictor $\othet$, and a \emph{bias term} $\la \| \othet\|^2$ which depends on the optimal predictor but not on the sample size $n$. All our bounds will have this form but with smaller quantities (but asking fore more assumptions).
Without further assumptions, in \cref{FR}, $\la$ is taken proportional to $\ {1}/{\sqrt{n}}$, and we get the usual optimal  slow rate in excess risk of $O(1/\sqrt{n})$ associated with such a general set-up~\citep[see, e.g.,][]{cesa2015complexity}.

For the specific case of quadratic losses of the form $\ell_z(\theta) = \frac{1}{2} ( y - \theta \cdot \Phi(x))^2$, where $z = (x,y)$, and $y \in \mathbb{R}$ and $\Phi(x) \in \mathcal{H}$, the situation is much richer. Without further assumptions, the same rate $O(1/\sqrt{n})$ is achieved, but stronger assumptions lead to faster rates~\citep{devito}. In particular, the decay of the eigenvalues of the Hessian $\E \big[ \Phi(x) \otimes \Phi(x) \big]$ (often called the \emph{capacity condition}) leads to an improved variance term, while the finiteness of some bounds on $\othet$ for norms other than the plain Hilbertian norms $\|\othet\|$ (often called the \emph{source condition}) leads to an improved bias term.
Both of these assumptions lead to faster rates than $O(1/\sqrt{n})$ for the excess risk, with the proper choice of the regularization parameter $\la$. For least-squares, these rates are then optimal and provide a better understanding of properties of the problem that influence the generalization capabilities of regularized ERM~\citep[see, e.g.][]{smale2007learning,devito,steinwart2009optimal,fischer2017sobolev,blanchard2018optimal}.

Our  main goal in this paper is to bridge the gap between Lipschitz-continuous   and quadratic losses by improving on slow rates for general classes of losses beyond least-squares. We first note that: (a) there has to be an extra regularity assumption because of lower bounds~\citep{cesa2015complexity}, and (b) asymptotically, we should obtain bounds that approach the local quadratic approximation of $\ell_z(\theta)$ around $\othet$ with the same optimal behavior as for plain least-squares.

Several frameworks are available for such an extension with extra assumptions on the losses, such as ``exp-concavity''~\citep{koren2015fast,mehta2016fast}, strong convexity~\citep{van2008high} or a generalized notion of self-concordance~\citep{bach2010self,ostrovskii2018finite}.
In this paper, we focus on self-concordance, which links the second and third order derivatives of the loss. This notion is quite general and corresponds to widely used losses in machine learning, and does not suffer from constants which can be exponential in  problem parameters (e.g., $\|\othet\|$) when applied to generalized linear models like logistic regression. See \cref{related} for a comparison to related work.

With this self-concordance assumption, we will show that our problem behaves like a quadratic problem corresponding to the local approximation around $\othet$, in a totally non-asymptotic way, which is the core technical contribution of this paper.
As we have already mentioned, this phenomenon is naturally expected in the asymptotic regime, but is hard to capture in the non-asymptotic setting without constants which explode exponentially with the problem parameters. 

The paper is organized as follows:
in \cref{main}, we present our main assumptions and  informal results, as well as  our bias-variance decomposition. In order to introduce precise results gradually, we start in \cref{simple} with a result similar to \cref{FR} for our set-up to show that we recover with a simple argument the result from~\citet{FR}, which itself applies more generally. Then, in \cref{bias} we introduce the source condition allowing for a better control of the bias. Finally, in \cref{sec:fast-rates-source-capacity}, we detail the capacity condition leading to an improved variance term, which, together with the improved bias leads to fast rates (which are optimal for least-squares).
\subsection{Related work}
\label{related}
\paragraph{Fast rates for empirical risk minimization.}
Rates faster than $O(1/\sqrt{n})$ can be obtained with a variety of added assumptions, such as some form of strong convexity~\citep{FR,boucheron2011high}, noise conditions for classification~\citep{steinwart2007fast}, or extra conditions on the loss, such as self-concordance~\citep{bach2010self} or exp-concavity~\citep{koren2015fast,mehta2016fast}, whose partial goal is to avoid exponential constants. Note that 
\citet{bach2010self} already considers logistic regression with Hilbert spaces, but only for well-specified models and a fixed design, and without the sharp and simpler results that we obtain in this paper.
\paragraph{Avoiding exponential constants for logistic regression.}
The problem of exponential constants (i.e., leading factors in the rates scaling as~$e^{RD}$ where~$D$ is the radius of the optimal predictor, and $R$ the radius of the design) is long known.
In fact,~\citet{hazan2014logistic} showed a lower bound, explicitly constructing an adversarial distribution (i.e., an ill-specified model) for which the problem manifests in the finite-sample regime with $n = O(e^{RD})$. Various attempts to address this problem are found in the literature. For example,~\citet[App.~C]{ostrovskii2018finite} prove the optimal $d/n$ rate in the non-regularized $d$-dimensional setting but, multiplied with the curvature parameter~$\rho$ which is at worst exponential but is shown to grow at most as~$(RD)^{3/2}$ in the case of Gaussian design.  Another approach is due to~\citet{foster2018logistic}: they establish ``$1$-mixability'' of the logistic loss, then apply Vovk's aggregating algorithm in the online setting, and then proceed via online-to-batch conversion. While this result allows to obtain the fast $O(d/n)$ rate (and its counterparts in the nonparametric setting) without exponential constants, the resulting algorithm is \textit{improper} (i.e., the canonical parameter~$\eta = \Phi(x)\cdot\theta^\star$, see below, is estimated by a \textit{non-linear} functional of $\Phi(x)$). 

A closely related approach is to use the notion of exp-concavity instead of mixability ~\citep{rakhlin2015online,koren2015fast,mehta2016fast}. The two close notions are summarized in the so-called central condition (due to~\citet{van2015fast}) which fully characterizes when the fast~$O(d/n)$ rates (up to log factors and in high probability) are available for improper algorithms. However, when proper learning algorithms are concerned, this analysis requires~$\eta$-mixability (or~$\eta$-exp-concavity) of the \textit{overall} loss~$\ell_{z}(\theta)$ for which the~$\eta$ parameter scales with the radius of the set of predictors. This scaling is exponential for the logistic loss, leading to exponential constants.

\section{Main Assumptions and Results}
\label{main}
Let $\Z$ be a Polish space and $Z$ be a random variable on $\Z$ with distribution $\rho$. 
Let $\hh$ be a separable (non-necessarily finite-dimensional) Hilbert space, with norm $\|\cdot\|$, and let $\ell: \Z \times \hh \to \R$ be a loss function, we denote by $\ell_z(\cdot)$ the function $\ell(z, \cdot)$.
Our goal is to minimize the expected risk with respect to $\theta \in \hh$:
\vspace{-2pt}
$$ \inf_{\theta \in \hh}~\Lo(\theta) = \Exp{\ell_Z(\theta)}.$$ 
Given $(z_i)_{i=1}^n \in \Z^n$, we will consider the following estimator based on regularized empirical risk minimization given $\la>0$ (note that the minimizer is unique in this case):
\vspace{-4pt}
\[
\othetnl = \argmin_{\theta \in \hh}\ \  \frac{1}{n}\sum_{i=1}^n \ell_{z_i}(\theta)  + \frac{\la}{2} \|\theta\|^2,
\vspace{-5pt}
\]
where we assume the following.
\begin{restatable}[i.i.d. data]{ass}{asmiid}
\label{asm:iid}
The samples $(z_i)_{1 \leq i \leq n}$ are independently and identically distributed according to $\rho$. 
\end{restatable}
The goal of this work is to provide upper bounds in high probability for the so-called {\em excess risk} \vspace{-5pt}
$$ L(\othetnl) - \inf_{\theta \in \hh} \ L(\theta),\vspace*{-5pt}$$
 and thus to provide a general framework to measure the quality of the estimator $\othetnl$. Algorithms for obtaining such estimators have been extensively studied, in both finite-dimensional regimes, where a direct optimization over $\theta$ is performed, typically by gradient descent or stochastic versions thereof~\citep[see, e.g.,][]{bottou2008tradeoffs,shalev2011pegasos} and infinite-dimensional regimes, where kernel-based methods are traditionally used~\citep[see, e.g.,][and references therein]{keerthi2005fast,gerfo2008spectral,dieuleveut2016nonparametric,tu2016large,rudi2017falkon}. 
\begin{example}[Supervised learning]\label[example]{examples:losses definitions}
Although formulated as a general $M$-estimation problem~\citep[see, e.g.,][]{lehmann2006theory}, our main motivation comes from supervised learning, with $\Z = \X \times \Y$ where $\X$ is the data space and $\Y$ the target space. We will consider, as examples, losses with both real-valued outputs but also  the multivariate case.
For learning real-valued outputs, consider we have a bounded representation of the input space $\Phi : \X \rightarrow \hh$ \citep[potentially implicit when using kernel-based methods,][]{aronszajn1950theory}. We will provide bounds for the following losses. 
\begin{itemize}
    \item The square loss $\ell_z(\theta) = \frac{1}{2}\left(y - \theta \cdot \Phi(x)\right)^2$, which is not Lipschitz-continuous.
    \item The Huber losses $\ell_z(\theta) = \psi(y - \theta \cdot \Phi(x))$ where $\psi(t) = \sqrt{1 + t^2} -1$ or $\psi(t) = \log \frac{e^t + e^{-t}}{2}$ \citep{hampel2011robust}, which are Lipschitz-continuous. 
   \item The logistic loss $\ell_z(\theta) = \log(1 + e^{- y \theta \cdot \Phi(x)})$ commonly used in binary classification where $y \in \{-1,1\}$, which is Lipschitz-continuous.
\end{itemize}
Our framework goes beyond real-valued outputs, and can be applied to all generalized linear models (GLM)~\citep{mccullagh1989generalized}, including softmax regression: we consider a representation function $\Phi : \X \times \Y \rightarrow \hh$ and an a priori measure $\mu$ on $\Y$. The loss we consider in this case is
\vspace*{-5pt}$$ \textstyle \ell_z(\theta) = -\theta \cdot \Phi(x,y) + \log \int_{\Y}{\exp\left(\theta \cdot \Phi(x,y^{\prime})\right) d\mu( y^{\prime})},$$
which corresponds to the negative conditional log-likelihood when modelling $y$ given $x$ by the distribution $p(y|x,\theta) \sim \frac{\exp\left(\theta \cdot \Phi(x,y)\right)}{\int_{\Y}{\exp\left(\theta \cdot \Phi(x,y^{\prime})\right) d\mu( y^{\prime})}}  d\mu(y) $. Our framework applies to all of these generalized linear models with almost surely bounded features $\Phi(x,y)$, such as conditional random fields~\citep{laffertyconditional}.
\end{example}
We can now introduce the main technical assumption on the loss $\ell$.
\begin{restatable}[Generalized self-concordance]{ass}{asmgensc}
\label{asm:gen_sc}
For any $z \in \Z$, the function $\ell_z(\cdot)$ is convex and three times differentiable. Moreover, there exists a set $\varphi(z) \subset \hh$ such that it holds \vspace{-7pt}:
\[\vspace*{-10pt}\forall \theta \in \hh,~\forall h,k \in \hh,~\left|\nabla^3 \ell_z(\theta)[k,h,h]\right| \leq \sup_{ g \in \varphi(z)}  |k \cdot g|  \ ~\nabla^2 \ell_z(\theta)[h,h]. \]
\end{restatable}
This is a generalization of the assumptions introduced by~\citet{bach2010self}, by allowing a varying term $\sup_{ g \in \varphi(z)}  |k \cdot g|$ instead of a uniform bound proportional to  $\| k \| $. This is crucial for the fast rates we want to show.  
\begin{example}[Checking assumptions]\label{ex:checkingasm}
For the losses in \cref{examples:losses definitions}, this condition is satisfied with the following corresponding set-function~$\varphi$. 
\begin{itemize}
    \item For the square loss $\ell_z(\theta) = \frac{1}{2}\left(y - \theta \cdot \Phi(x)\right)^2$, $\varphi(z) = \{0\}$.
    \item For the Huber losses $\ell_z(\theta) = \psi(y - \theta \cdot \Phi(x))$, if $\psi(t) = \sqrt{1 + t^2} -1$, then $\varphi(z) = \{  3\Phi(x) \} $ and if $\psi(t) = \log \frac{e^t + e^{-t}}{2}$, then $\varphi(z) = \{ 2\Phi(x) \}$~\citep{ostrovskii2018finite}. For the logistic loss $\ell_z(\theta) = \log(1 + e^{- y \theta \cdot \Phi(x)})$, we have $\varphi(z) = \{ y \Phi(x) \}$ (here, $\varphi(z)$ is reduced to a point). 
   \item For generalized linear models, 
   $\nabla^3 \ell_z(\theta)$ is a third-order cumulant, and thus
   $\left|\nabla^3 \ell_z(\theta)[k,h,h]\right| \leq
   \E_{p(y|x,\theta)} | k \cdot \Phi(x,y) - k \cdot \E_{p(y'|x,\theta)}\Phi(x,y') |\cdot 
   | h \cdot \Phi(x,y) - h \cdot \E_{p(y'|x,\theta)}\Phi(x,y') |^2
   \leq 2 \sup_{y \in \Y} | k \cdot \Phi(x,y) | \   \nabla^2 \ell_z(\theta)[h,h] $. Therefore
   $\varphi(z) = \{ 2\Phi(x,y'), \ y'
   \in \Y\}$ (which is not a singleton).
\end{itemize}
\end{example}
Moreover we require the following two technical assumptions to guarantee that $L(\theta)$ and its first and second derivatives are well defined for any $\theta \in \hh$.
\begin{restatable}[Boundedness]{ass}{asmbounded}
\label{asm:bounded}
There exists $\Rad \geq 0$ such that $\sup_{ g \in \varphi(Z) } \| g \| \leq \Rad$ almost surely. 
\end{restatable}

\vspace*{-.6cm}

\begin{restatable}[Definition in $0$]{ass}{asmwd}
\label{asm:wd} 
$\left|\ell_Z(0)\right|$, $\|\nabla \ell_Z(0)\|$ and  $\Tr(\nabla^2 \ell_Z(0))$ are almost surely bounded.
\end{restatable}
The assumptions above are usually easy to check in practice. In particular, if the support of~$\rho$ is bounded, the mappings $z \mapsto \ell_z(0),\nabla \ell_z(0), \Tr(\nabla^2\ell_z(0))$ are continuous, and $\varphi$ is uniformly bounded on bounded sets, then they hold. The main regularity assumption we make on our statistical problems follows. 
\begin{restatable}[Existence of a minimizer]{ass}{asmminimizer}
\label{asm:minimizer}
There exists $\othet \in \hh$ such that $ \Lo(\othet) = \inf_{\theta \in \hh} \Lo(\theta).$
\end{restatable}
While \cref{asm:bounded} is standard in the analysis of such models~\citep{devito,FR,steinwart2009optimal,bach2014adaptivity}, \cref{asm:minimizer} imposes that the model is ``well-specified'', that is, for supervised learning situations from \cref{examples:losses definitions}, we have chosen a rich enough representation $\Phi$. It is possible to study the non-realizable case in our setting by requiring additional technical assumptions (see \cite{steinwart2009optimal} or discussion after \eqref{radius}), but this is out of scope of this paper. Note that our well-specified assumption (for logistic regression for simplicity of arguments) is weaker than requiring $f^\star(x) = \Exp{Y|X}$ being equal to $\theta^\star \cdot \Phi(x)$.
We can now introduce the main definitions allowing our bias-variance decomposition.
\bd[Hessian, Bias, Degrees of freedom]
Let $\Ll(\theta) = \L(\theta) \!+\! \frac{\la}{2} \|\theta\|^2$; define 
the \emph{expected Hessian} $\Ho(\theta)$, the regularized Hessian $\Hl(\theta)$, the \emph{bias} $\biass_\la$ and the \emph{degrees of freedom} $\Deff_\la$~as: 
\vspace{-2pt}
\eqal{
\Ho(\theta) &= \Exp{\nabla^2 \ell_Z(\theta)}, \ \mbox{ and } \ 
\Hl(\theta) =   \Ho(\theta) + \lambda I, \\
\biass_\la &= \|\Hl(\othet)^{-1/2} \nabla \Ll(\othet)\|,\\
\Deff_\la &= \Exp{\|\Hl(\othet)^{-1/2}\nabla \ell_Z(\othet)\|^2}.
\vspace{-2pt}
}
\ed
Note that the bias and degrees of freedom only depend on the optimum $\othet \in \hh$ and not on the minimizer $\othetl$ of the regularized expected risk.
Moreover, the degrees of freedom $\Deff_\la$ correspond to the usual  Fisher information term commonly seen in the asymptotic analysis of $M$-estimation~\citep{van2000asymptotic,lehmann2006theory}, and correspond to the usual quantities introduced in the analysis of least-squares~\citep{devito}. Indeed, in the least-squares case, we recover exactly 
$ \biass_\la = \la\|\mathbf{C}_\la^{-1/2} \othet\|$ and $\Deff_\la = \Tr(\mathbf{C} \mathbf{C}_\la^{-1}),$
where $\mathbf{C}$ is the covariance operator $\mathbf{C} = \Exp{\Phi(x) \otimes \Phi(x)}$ and $\mathbf{C}_\la = \mathbf{C} + \la I$.

Our results will rely on the quadratic approximation of the losses around $\othet$. Borrowing tools from the analysis of Newton's method~\citep{nesterov1994interior}, this will only be possible in the vicinity of $\othet$. The proper notion of vicinity is the so-called  {\em radius of the Dikin ellipsoid}, which we define as follows: 
\vspace{-2pt}
\eqal{
\label{radius}
\rla(\theta) & \quad \textrm{such that}\quad 1/\rla(\theta) = \sup_{z \in \supp(\rho)} \sup_{g \in \varphi(z)} \|\Hl^{-1/2}(\theta)g\|. 
\vspace{-6pt}
}
 Our most refined bounds will depend whether the bias term is small enough compared to $\rla(\othet)$. We believe that in the non realizable setting, the results we obtain would still hold when the bias term is smaller than the Dikin radius, although one would have to modify the definitions to incorporate the fact that $\theta^\star$ is not in $\hh$.  The following informal result summarizes all of our results.
\bt[General bound, informal]
Let $n \in \N$, $\delta \in (0,1/2]$, $\la > 0$. Under \cref{asm:gen_sc,asm:wd,asm:bounded,asm:iid,asm:minimizer}, whenever
\vspace{-4pt}
\[n \geq C_0 \frac{\Rad^2 \Deff_\la ~\log \frac{2}{\delta}}{\la},
\]
then with probability at least $1-2 \delta$, it holds
\vspace{-4pt}
$$\L(\othetnl) - \L(\othet) \leq \mathsf{C}_{\textup{bias}}~\biass_\la^2 + \mathsf{C}_{\textup{var}}~\frac{\Deff_\la~\log \frac{2}{\delta}}{n},
\vspace{-2pt}
$$
where $C_0, \mathsf{C}_{\textup{bias}}$ and $\mathsf{C}_{\textup{var}}$ are either universal or depend only on $\Rad \|\othet\|$. 
\et
This mimics a usual bias-variance decomposition, with a bias term $\biass_\la^2$ and a variance term  proportional to ${\Deff_\la}/{n}$. In particular in the rest of the paper we quantify the constants and the rates under various regularity assumptions, and specify the good choices of the regularization parameter~$\la$.
In \cref{tab:summary}, we summarize the different assumptions and corresponding rates.

\begin{table}[]
    \centering
    \begin{tabular}{c|ccccl}
\toprule
  Assumptions & Bias & Variance & Optimal $\lambda$ & Optimal Rate &  \\
  \midrule
  None & $\lambda$ & $\frac{1}{\lambda n}$ & $n^{-1/2}$ & $n^{-1/2}$ & \cref{thm:gen_bound_222,cor:base-rates} \\
  Source
  & $\lambda^{2r+1}$ & $\frac{1}{\lambda n}$ & $n^{-\frac{1}{2r+2}}$ & $n^{-\frac{2r+1}{2r+2}}$ & \cref{thm:general-result-weak,cor:bounds_source}  \\
    Source + Capacity
  & $\lambda^{2r+1}$ & $\frac{1}{
  \lambda^{1/\alpha} n}$ & $n^{-\frac{\alpha}{2r\alpha+\alpha+1}}$ & $n^{-\frac{2r\alpha+\alpha}{2r\alpha+\alpha+1}}$ & \cref{thm:general-result,cor:bounds_source_capacity}  \\
\bottomrule
    \end{tabular}
    \caption{Summary of convergence rates, without constants except $\lambda$, for source condition (Asm.~\ref{asm:source-condition}): $\othet \in \textrm{Im}(\Ho(\othet)^r)$,~  $r \in (0,1/2]$, capacity condition (Asm.~\ref{asm:capacity}): $\Deff_\la = O(\la^{-1/\alpha})$,~ $\alpha \geq 1$ .} 
    \label{tab:summary}
    \vspace*{-20pt}
\end{table}

\section{Slow convergence rates}
\label{simple}
Here we bound the quantity of interest without any regularity assumption (e.g., source of capacity condition) beyond some boundedness assumptions on the learning problem. We consider the various bounds on the derivatives of the loss $\ell$:
\vspace{-4pt}
$$\Bone(\theta) = \!\!\!\sup_{z \in \supp(\rho)} \!\! \|\nabla \ell_z(\theta)\|, \  \Btwo(\theta) = \!\!\!\sup_{z \in \supp(\rho)}\!\! \Tr(\nabla^2 \ell_z(\theta)), \ \Bob= \!\!\sup_{\|\theta\| \leq \|\othet\|} \!\Bone(\theta)   , \  \Btb = \!\!\sup_{\|\theta\|\leq \|\othet\|}\! \Btwo(\theta).$$

\begin{example}[Bounded derivatives]
In all the losses considered above, assume the feature representation ($\Phi(x)$ for the Huber losses and the square loss, $y\Phi(x)$ for the logistic loss, and $\Phi(x,y)$ for GLMs) is bounded by $\bar{R}$. Then the losses considered above apart from the square loss are Lipschitz-continuous and $\Bone$ is uniformly bounded by $\bar{R}$. For these losses, $\Btwo$ is also uniformly bounded by~$\bar{R}^2$. Using \cref{ex:checkingasm}, one can take $\bar{R}$ to be equal to a constant times $R$ ($1/2$ and $1/3$ for the respective Huber losses, $1$ for logistic regression and $1/2$ for canonical GLMs).
 For the square loss (where $R=0$ because the third-order derivative is zero), 
$\Btb \leq \bar{R}^2$ and $\Bob \leq \bar{R} \| y\|_\infty + \bar{R}^2 \|\othet\|$, where $\| y\|_\infty$ is an almost sure bound on the output $y$.
\end{example}
\bt[Basic result]\label[theorem]{thm:gen_bound_222}
Let $n \in \N$ and $0 < \la \leq \Btb$. Let $\delta \in (0,1/2]$. If 
\vspace{-5pt}
\[ n \geq 512 \left( \|\othet\|^2 \Rad^2\vee 1\right) \log \frac{2}{\delta},~~~n\geq 24 \frac{\Btb}{\la} \log \frac{8 \Btb}{\la \delta}, ~~~ n \geq 256 \frac{\Rad^2\Bob^2}{ \la^2 } ~\log \frac{2}{\delta}, \]
then with probability at least $1 - 2 \delta$, 
\vspace{-10pt}
\eqal{
L(\othetnl) - L(\othet) \leq 84~ \frac{\Bob^2}{\la n}~\log \frac{2}{\delta} +2 \lambda \|\othet\|^2.}
\et
This result shown in \cref{sec:main-simple-final} as a consequence of \cref{thm:gen_bound_22} (also see the proof sketch in \cref{sec:proof}) matches the one obtained with Lipschitz-continuous losses~\citep{FR} and  the one for least-squares when assuming the existence of $\othet$~\citep{devito}. The following corollary (proved as \cref{thm:quantitative_slow_rates} in  \cref{sec:explicit-simple}) gives the bound optimized in~$\la$, with explicit rates. 
\bcor[Basic Rates]\label[corollary]{cor:base-rates}
Let $\delta \in (0,1/2]$. Under \cref{asm:iid,asm:gen_sc,asm:bounded,asm:wd,asm:minimizer}, when $n \geq N, \la = C_0\sqrt{\log(2/\delta)/n}$, then with probability at least $1-2\delta$,
\vspace{-4pt}
$$ L(\othetnl) - L(\othet) \leq  C_1 ~ n^{-1/2} ~ \log^{1/2}  \frac{2}{\delta}.
\vspace{-4pt}
$$
 with $C_0 =16 \Bob \max(1,\Rad),C_1 =  48 \Bob \max(1,\Rad)\max(1,\|\othet\|^2)$ and with $N$ defined in \cref{eq:constantNslow} and satisfying $N \!\!=\!\!  O(\poly(\Bob,\Btb,\Rad \|\othet\|))$ where $\poly$ denotes a polynomial function of the inputs.
\ecor
Both bias and variance terms are of order $O(1/\sqrt{n})$ and we recover up to constants terms the result of~\citet{FR}. In the next section, we will improve both bias and variance terms to obtain faster rates.
\section{Faster Rates with Source Conditions}
\label{bias}
Here we provide a more refined bound, where we introduce a \emph{source condition} on $\othet$ allowing to improve the bias term and to achieve learning rates as fast as $O(n^{-2/3})$.
We first define the localized versions of $\Bob, \Btb$:
\vspace{-2pt}
$$\Bos = \Bone(\othet), \qquad \Bts = \Btwo(\othet),
\vspace{-2pt}$$
and recall the definition of the bias 
\vspace{-4pt}
\eqal{
\label{eq:bias-def}
\biass_\la &= \|\Hl(\othet)^{-1/2} \nabla \Ll(\othet)\|. 
\vspace{-4pt}}
Note that since $\othet$ is the minimizer of $\L$, we have $\nabla \L(\othet) = 0$, so that
$\nabla \Ll(\othet) = \nabla \L(\othet) + \la \othet = \la \othet$,
and  $\biass_\la = \la \|\Hl(\othet)^{-1/2} \othet\|.$
This characterization is always bounded by $\la \|\othet\|^2$, but allows a finer control of the regularity of $\othet$, leading to improved rates compared to \cref{simple}.

Note that in the least-squares case, we recover exactly the bias of ridge regression
 $\biass_\la = \la\|\mathbf{C}_\la^{-1/2} \othet\|$,
where $\mathbf{C}$ is the covariance operator $\mathbf{C} = \Exp{\Phi(x) \otimes \Phi(x)}$.

Using self-concordance, we will relate quantities at $\othet$ to quantities at $\othetl$ using:
\vspace{-2pt}
$$\tla = \sup_{z \in \supp(\rho)} \sup_{g \in \varphi(z)} |(\othetl - \othet) \cdot g|.
\vspace{-2pt}
$$
The following theorem, proved in \cref{sec:final_results_general}, relates $\biass_\la$ to the excess risk.
\bt[Decomposition with refined bias]\label[theorem]{thm:general-result-weak}
Let $n \in \N$, $\delta \in (0,1/2]$, $0 < \la \leq \Bts$. Whenever
\[n \geq \triangle_1 \frac{\Bts}{\la}\log \frac{8 \square_1^2\Bts}{\la \delta},~~~n \geq \triangle_2~  \frac{(\Bos \Rad)^2}{\la^2} ~\log \frac{2}{\delta},\]
 then with probability at least $1-2 \delta$, it holds
\eqal{\label{eq:general-result-weak}
\L(\othetnl) - \L(\othet) \leq \mathsf{C}_{\textup{bias}}~\biass_\la^2 + \mathsf{C}_{\textup{var}}~\frac{(\Bos)^2}{\la n}~\log \frac{2}{\delta},
\vspace{-2pt}
}
where $\square_1 \leq e^{\tla/2},\triangle_1 \leq 2304 e^{4 \tla} (1/2 \vee \Rad \|\othet\|),\triangle_2 \leq 256 e^{2 \tla}, \Cb \leq 6 e^{2 \tla}, \Cv \leq 256 e^{3 \tla}$.
\et

It turns out that the  {\em radius of the Dikin ellipsoid} $\rla(\othet)$ defined in \cref{radius} provides the sufficient control over the constants above: when the bias is of the same order of the radius of the Dikin ellipsoid, the quantities $\mathsf{C}_{\textup{bias}},\mathsf{C}_{\textup{var}},\triangle_1,\triangle_2$ become universal constants instead of depending exponentially on $R \| \othet \|$, as shown by the lemma below, proved in \cref{lm:tla} in \cref{sec:app-main-result}.
\blm\label[lemma]{lm:bound-tla-main}
When $\biass_\la \leq \frac{\rla(\othet)}{2}$ then $\tla \leq \log 2$ else $\tla \leq 2R\|\othet\|$.
\elm
Interestingly, regularity of $\othet$, like the source condition below, can induce this effect, allowing a better dependence on $\la$ for the bias term.
\ba[Source condition]\label{asm:source-condition}
There exists $r \in (0,1/2]$ and $v \in \hh$ such that
$\othet = \Ho(\othet)^r v$.
\ea
In particular we denote by $\Lc := \|v\|$. \cref{asm:source-condition} is commonly made in least-squares regression~\citep{devito,steinwart2009optimal,blanchard2018optimal} and is equivalent to requiring that, when expressing $\othet$ with respect to the eigenbasis of $\Ho(\othet)$, i.e., $\othet = \sum_{j \in \N} \alpha_j u_j$,  where $\la_j, u_j$ is the eigendecomposition of $\Ho(\othet)$, and $\alpha_j = \theta \cdot u_j$, then $\alpha_j$ decays as $\la_j^r$. In particular, with this assumption, defining $\beta_j = v \cdot u_j$,
\vspace{-4pt}
$$\biass_\la^2 = \la^2 \sum_{j} \frac{\alpha_j^2}{\la_j + \la} = \la^2 \sum_{j} \frac{\la_j^{2r}\beta_j^2}{\la_j + \la} \leq \la^2 \Big(\sup_{j} \frac{\la_j^{2r}}{\la_j + \la}\Big) \sum_j \beta_j^2 \leq \la^{1+2r} \|v\|^2.
\vspace{-4pt}
$$
Note moreover that $\Ho(\othet) \preccurlyeq \Bts \mathbf{C}$, meaning that the usual sufficient conditions leading to the source conditions for least-squares also apply here. For example, for logistic regression, if the log-odds ratio is smooth enough, then it is in $\hh$. So, when $\hh$ corresponds to a Sobolev space of smoothness $m$ and the marginal of $\rho$ on the input space is a density bounded away from $0$ and infinity with bounded support, then the source condition corresponds essentially to requiring $\othet$ to be $(1+2r)m$-times differentiable \citep[see discussion after Thm. 9 of][for more details]{steinwart2009optimal}. A precise example can be found in Sec. 4.1 of \cite{Pillaud-Vivien18}. 

In conclusion, the effect of additional regularity for $\othet$ as \cref{asm:source-condition}, has two beneficial effects: (a) on one side it allows to obtain faster rates as shown in the next corollary, (b) as mentioned before, somewhat surprisingly, it reduces the constants to universal, since it allows the bias to go to zero faster than the Dikin radius (indeed, the squared radius $\rla^2(\othet)$ is always larger than $\lambda / R^2$, which is strictly larger than $\lambda^{1+2r} \|v\|^2$ if $r>0$ and $\lambda$ small enough). This is why we do not the get exponential constants imposed by~\citet{hazan2014logistic}.
\bcor[Rates with source condition]\label[corollary]{cor:bounds_source}
Let $\delta \in (0,1/2]$. Under \cref{asm:iid,asm:wd,asm:bounded,asm:gen_sc,asm:minimizer} and \cref{asm:source-condition}, whenever $n \geq N$ and
$\lambda = (C_0/n)^{1/(2+2r)}$, then
     with probability at least $1-2 \delta$,  
    \[\Lo(\othetnl) - \Lo(\othet) ~~~\leq~~~  C_1~~n^{-\frac{1 + 2r}{2+2r}} \log \frac{2}{\delta},\]
with $C_0 = 256~ (\Bos/\Lc)^{2},~C_1 = 8~(256)^{\gamma}((\Bos)^{\gamma} \Lc^{1-\gamma})^2, \gamma = \frac{1 + 2r}{2+2r}$ and with $N$ defined in \cref{defNr} and satisfying $N = O(\poly(\Bos,\Bts,\Lc,\Rad,\log(1/\delta)))$. 
\ecor
The corollary above, derived in \cref{sec:explicit-general}, is obtained by minimizing in $\la$ the r.h.s. side of \cref{eq:general-result-weak} in \cref{thm:general-result-weak}, and considering that when $\othet$ satisfies the source condition, then $\biass_\la \leq \la^{1+2r} \Lc$, while the variance is still of the form ${1}/{(\la n)}$. When $r$ is close to $0$, the rate $1/\sqrt{n}$ is recovered. When instead the target function is more regular, implying $r=1/2$, a rate of $n^{-2/3}$ is achieved. Two considerations are in order: (a) the obtained rate is the same as least-squares and minimax optimal \citep{devito,steinwart2009optimal,blanchard2018optimal}, (b) the fact that regularized ERM is adaptive to the regularity of the function up to $r = 1/2$ is a byproduct of Tikhonov regularization as already shown for the least-squares case by \citet{gerfo2008spectral}. Using different regularization techniques may remove the limit $r=1/2$.

\section{Fast Rates with both Source and Capacity Conditions}
\label{sec:fast-rates-source-capacity}
In this section, we consider improved results with a finer control of the effective dimension $\Deff_\la$ (often called degrees of freedom), which, together with the source condition allows to achieve rates as fast as $1/n$:
\vspace{-10pt}
\eqals{
\Deff_\la &= \Exp{\|\Hl(\othet)^{-1/2}\nabla \ell_Z(\othet)\|^2},
\vspace{-4pt}
}
As mentioned earlier this definition of $\Deff_\la$ corresponds to the usual asymptotic term in $M$-estimation.
Moreover, in the case of least-squares, it corresponds to the standard notion of effective dimension $\Deff_\la = \Tr(\mathbf{C} \mathbf{C}_\la^{-1})$ \citep{devito,blanchard2018optimal}. Note that by definition, we always have $\Deff_\la \leqslant \Bos{}^2/\la$, but we can have in general a much finer control. For example, for least-squares, $\Deff_\la = O(\la^{-1/\alpha})$ if the eigenvalues of the covariance operator $\mathbf{C}$ decay as $\la_j(\mathbf{C}) = O( j^{-\alpha})$, for $\alpha \geq 1$. Moreover note that since $\mathbf{C}$ is trace-class, by Asm.~\ref{asm:bounded}, the eigenvalues form a summable sequence and so $\mathbf{C}$ satisfies $\la_j(\mathbf{C}) = O( j^{-\alpha})$ with $\alpha$ always larger than $1$. 
\begin{example}[Generalized linear models] For generalized linear models, an extra assumption makes the degrees of freedom particularly simple: if the probabilistic model is well-specified, that is, there exists $\othet$ such that almost
surely, $p(y|x) = p(y|x,\othet) = \frac{\exp\left(\othet \cdot \Phi(x,y)\right)}{\int_{\Y}{\exp\left(\othet \cdot \Phi(x,y^{\prime})\right) d\mu( y^{\prime})}} $, then from the
usual Bartlett identities~\citep{bartlett1953approximate} relating the expected squared derivatives and Hessians, we have
$\Exp{ \nabla \ell_z(\othet) \otimes \nabla \ell_z(\othet) } = \Ho(\othet)$, leading to
$\Deff_\la =  \Tr(\Hl(\othet)^{-1} \Ho(\othet))$. 
\end{example}
As we have seen in the previous example there are interesting problems for which $\Deff_\la = \Tr (\Ho(\othet) + \lambda I)^{-1} \Ho(\othet))$. Since we have $\Ho(\othet) \preceq \Bts \mathbf{C}$, $\Deff_\la$ still enjoys a polynomial decay depending on the eigenvalue decay of $\mathbf{C}$ as observed for least-squares. In the finite-dimensional setting where $\hh$ is of dimension $d$, note that in this case, $\Deff_\la$ is always bounded by $d$. Now we are ready to state our result in the most general form, proved in \cref{sec:final_results_general}.
\bt[General bound]\label[theorem]{thm:general-result}
Let $n \in \N$, $\delta \in (0,1/2]$, $0 < \la \leq \Bts$. Whenever
\[n \geq \triangle_1 \frac{\Bts}{\la}\log \frac{8 \square_1^2\Bts}{\la \delta},~~~n \geq \triangle_2~  \frac{\Deff_\lambda \vee (\Qs)^2}{\rla(\othet)^2} ~\log \frac{2}{\delta},\]
with $(\Qs)^2 = \Bos{}^2/\Bts$, then with probability at least $1-2 \delta$, it holds
\eqal{\label{eq:general-result-bound}
\L(\othetnl) - \L(\othet) \leq \mathsf{C}_{\textup{bias}}~\biass_\la^2 + \mathsf{C}_{\textup{var}}~\frac{\Deff_\la \vee (\Qs)^2}{n}~\log \frac{2}{\delta},
}
where, $\mathsf{C}_{\textup{bias}},\mathsf{C}_{\textup{var}}, \square_1 \leq 414,~\triangle_1,~\triangle_2 \leq 5184$ ~when~ $\biass_\la \leq \rla(\othet)/2$; \\ otherwise $\mathsf{C}_{\textup{bias}},~\mathsf{C}_{\textup{var}},~\square_1 \leq 256 e^{6 \Rad \|\othet\|},~\triangle_1,~\triangle_2 \leq 2304 (1+\Rad\|\othet\|)^2e^{8\Rad\|\othet\|}$.
\et
As shown in the theorem above, the variance term depends on $\Deff_\la/n$, implying that, when $\Deff_\la$ has a better dependence in $\la$ than $1/\la$, it is possible to achieve faster rates. We quantify this with the following assumption. 
\ba[Capacity condition]\label{asm:capacity}
There exists $\alpha > 0$ and $\Qc \geq 0$ such that $\displaystyle \Deff_\la \leq \Qc \la^{-1/\alpha}$.
\ea
\cref{asm:capacity} is standard in the context of least-squares, \citep{devito} and in many interesting settings is implied by the eigenvalue decay order of $\Ho(\othet)$, or $\mathbf{C}$ as discussed above. In the following corollary we quantify the effect of $\Deff_\la$ in the learning rates.
\bcor\label[corollary]{cor:bounds_source_capacity}
Let $\delta \in (0,1/2]$. Under \cref{asm:gen_sc,asm:wd,asm:bounded,asm:iid,asm:minimizer}, \cref{asm:source-condition} and \cref{asm:capacity}, when $n \geq N$ and $\la = (C_0/n)^{\alpha/(1 + \alpha(1+2r))}$, then with probability at least $1-2\delta$,
\[\Lo(\othetnl) - \Lo(\othet) \leq C_1 n^{-\frac{\alpha(1+2r)}{1+\alpha(1+2r)}}\log \frac{2}{\delta},\]
with $C_0 = 256 (\Qc/\Lc)^2,~C_1 = 8 (256)^{\gamma}~(\Qc^{\gamma} ~ 
    \Lc^{1-\gamma})^2,~\gamma = \frac{\alpha(1+2r)}{1 + \alpha(1+2r)}$ and $N$ defined in \cref{defNr} and satisfying
$N = O(\poly(\Bos,\Bts,\Lc,\Qc,\Rad,\log(1/\delta)))$. 
\ecor
The result above is derived in \cref{cor:bounds_source_capacity_aux} in \cref{sec:explicit-general} and is obtained by bounding $\biass_\la$ with $\la^{1+2r} \Lc$ due to the source condition, and $\Deff_\la$ with $\la^{-1/\alpha}$ due to the capacity condition and then optimizing the r.h.s. of \cref{eq:general-result-bound} in $\la$. Note that (a) the learning rate under the considered assumptions is the same as least-squares and minimax optimal \citep{devito}, and (b) when $\alpha = 1$ the same rate of \cref{cor:bounds_source} is achieved, which can be as fast as $n^{-2/3}$, otherwise, when $\alpha \gg 1$, we achieve a learning rate in the order of $1/n$, for $\la = n^{-1/(1+2r)}$.
%

\section{Sketch of the proof}\label{sec:proof}
In this section we will use the notation $\|v\|_\mathbf{A} := \|\mathbf{A}^{1/2} v \|$, with $v \in \hh$ and $\mathbf{A}$ a bounded positive semi-definite operator on $\hh$.
Here we prove that the excess risk decomposes using the bias term $\biass_\la$ defined in \cref{eq:bias-def} and a variance term $V_\la$, where $V_\la$ is defined as
\vspace{-6pt}
$$V_\la := \|\nabla \Lnl(\othetl)\|_{\Hess_\la^{-1}(\othetl)}, ~~~\textrm{with}~~~ \Lnl(\cdot) = \frac{1}{n}\sum_{i=1}^n \ell_{z_i}(\cdot) + \frac{\la}{2} \|\cdot\|^2,
\vspace{-6pt}
$$
which in turn is a random variable that concentrate in high probability to $\sqrt{\Deff_\la/n}$.

\paragraph{Required tools.} To proceed with the proof we need two main tools. The first is a result on the equivalence of norms of the empirical Hessian $\Hnl(\theta) = \nabla^2\Lnl(\theta)$ w.r.t.~the true Hessian $\Hl(\theta) = \nabla^2\Ll(\theta)$ for $\la > 0$ and $\theta \in \hh$. The result is proven in \cref{lm:hessian_concentration} of \cref{sec:concentration_general}, using Bernstein inequalities for Hermitian operators \citep{tropp2012user}, and essentially states that for $\delta \in (0,1]$, whenever $n \geq \frac{24\Btwo(\theta)}{\la} \log \frac{8\Btwo(\theta)}{\la \delta}$, then with probability $1-\delta$, it holds
\eqal{\label{eq:change-norm-easily}
\|\cdot\|_{\Hl(\theta)} \leq 2 \|\cdot\|_{\Hnl(\theta)}, \qquad \|\cdot\|_{\Hnl^{-1}(\theta)} \leq 2 \|\cdot\|_{\Ho^{-1}_\la(\theta)}.
}
The second result is about localization properties induced by generalized self-concordance on the risk. We express the result with respect to a generic probability $\mu$ (we will use it with $\mu = \rho$ and $\mu = \frac{1}{n} \sum_{i=1}^n \delta_{z_i}$). Let $\mu$ be a probability distribution with support contained in the support of~$\rho$. Denote by $L_{\mu}(\theta)$ the risk $L_{\mu}(\theta) = \mathbb{E}_{z \sim \mu}[\ell_z(\theta)]$ and by $L_{\mu,\la}(\theta) = L_\mu(\theta) + \frac{\la}{2} \|\theta\|^2$ (then $L_{\mu,\la} = L_\la$ when $\mu = \rho$, or $\widehat{L}_\la$ when $\mu = \frac{1}{n} \sum_{i=1}^n \delta_{z_i}$).
\bp\label[proposition]{prop:self-concordance-main-text}
Under \cref{asm:gen_sc,asm:wd,asm:bounded},
the following holds: (a) $L_{\mu,\la}(\theta), \nabla L_{\mu,\la}(\theta), \Ho_{\mu,\la}(\theta)$ are defined for all  $\theta \in \hh, \la \geq 0$, (b) for all $\la > 0$, there exists a unique $\othet_{\mu, \la} \in \hh$ minimizing $L_{\mu,\la}$ over $\hh$, and (c) for all $\la > 0$ and $\theta \in \hh$, 
\eqal{\label{eq:hess_control_opt}
\Hess_{\mu,\la}(\theta)\preceq e^{\mathsf{t}_{0}}\Hess_{\mu,\la}(\othet_{\mu,\la}),\\
\label{eq:bound_to_opt}L_{\mu,\la}(\theta) - L_{\mu,\la}(\othet_{\mu,\la}) \leq \psi(\mathsf{t}_0)\| \theta - \othet_{\mu,\la} \|^2_{\Hess_{\mu,\la}(\othet_{\mu,\la})},\\
\label{eq:grad_inf_opt} \dikin(\mathsf{t}_{0}) \|\theta - \othet_{\mu,\la}\|_{\Hess_{\mu,\la}(\theta)} \leq \|\nabla L_{\mu,\la}(\theta)\|_{\Hess_{\mu,\la}^{-1}(\theta)},
}
(d) \cref{eq:hess_control_opt,eq:bound_to_opt} hold also for $\la = 0$, provided that $\othet_{\mu,0}$ exists. Here $\mathsf{t}_{0} := \nm{\theta- \othet_{\mu,\la}}$ and $ \dikin(t) = (1- e^{-t})/t, ~~\psi(t) = (e^t - t- 1)/t^2$.
\ep
The result above is proved in \cref{subsec:basicsc} and is essentially an extension of results by \cite{bach2010self} applied to $L_{\mu,\la}$ under \cref{asm:gen_sc,asm:wd,asm:bounded}. 

\paragraph{Sketch of the proof.} Now we are ready to decompose the excess risk using our bias and variance terms. In particular we will sketch the decomposition without studying the terms that lead to constants terms. For the complete proof of the decomposition see \cref{prp:anal_dec} in \cref{sec:analytic_bounds_general}. Since $\othet$ exists by \cref{asm:minimizer}, using \cref{eq:bound_to_opt}, applied with $\mu = \rho$ and $\la = 0$, we have $\Lo(\theta) - \Lo(\othet) \leq \psi(\nm{\theta-\othet})\|\theta - \othet\|^2_{\Ho(\othet)}$ for any $\theta \in \hh$. By setting $\theta = \othetnl$, we obtain
\vspace{-5pt}
$$\Lo(\othetnl) - \Lo(\othet) \leq \psi(\nm{\othetnl-\othet})\|\othetnl - \othet\|^2_{\Ho(\othet)}.
\vspace{-5pt}
$$
The term $\psi( \nm{\othetnl-\othet})$ will become a constant. For the sake of simplicity, in this sketch of proof we will not deal with it nor with other terms of the form $\nm{\cdot}$ leading to constants. On the other hand, the term $\|\othetnl - \othet\|^2_{\Ho(\othet)}$ will yield our bias and variance terms.
Using the fact that $\Ho(\othet)\preceq \Ho(\othet) + \la I =: \Hl(\othet)$, by adding and subtracting $\othetl$, we have
\vspace{-5pt}
$$\|\othetl - \othet\|_{\Ho(\othet)} \leq \|\othetl - \othet\|_{\Hl(\othet)} \leq \|\othetl - \othet\|_{\Hl(\othet)} + \|\othetnl - \othetl\|_{\Hl(\othet)},
\vspace{-5pt}
$$
so
\vspace{-5pt}
$$
\Lo(\othetnl) - \Lo(\othet) ~~\leq~~ \textrm{const.}~\times~(\|\othetl - \othet\|_{\Hl(\othet)}^2 + \|\othetnl - \othetl\|_{\Hl(\othet)})^2.
\vspace{-5pt}
$$
By applying \cref{eq:hess_control_opt} with $\mu = \rho$ and $\theta = \othet$,  we have $\Hl(\othet) \preceq e^{\tla} \Hl(\othetl)$ and so we further bound $\|\othetnl - \othetl\|_{\Hl(\othet)}$ with $e^{\tla/2}\|\othetnl - \othetl\|_{\Hl(\othetl)}$ obtaining
\eqals{
\begin{aligned}
\Lo(\othetnl) - \Lo(\othet) ~~\leq~~ \textrm{const.}~\times~(\|\othetl - \othet\|_{\Hl(\othet)} + e^{\tla/2}\|\othetnl - \othetl\|_{\Hl(\othetl)})^2.
\end{aligned}
}
The term $\|\othetl - \othet\|_{\Hl(\othet)}$ will lead to the \textit{bias terms}, while the term $\|\othetnl - \othetl\|_{\Hl(\othetl)}$ will lead to the \textit{variance term}.
\paragraph{Bounding the bias terms.}
Recall the definition of bias $\biass_\la = \|\nabla\Ll(\othet)\|_{\Hess^{-1}_{\la}(\othet)}$ and of the constant $\tla := \nm{\othet - \othetl}$.
We bound $\|\othet - \othetl\|_{\Hl(\othet)}$ by applying \cref{eq:grad_inf_opt} with $\mu = \rho$ and $\theta = \othet$
\[\|\othet - \othetl\|_{\Hl(\othet)}~~\leq~~ 1/\dikin(\tla)~ \|\nabla\Ll(\othet)\|_{\Hess^{-1}_{\la}(\othet)} ~~=~~ 1/\dikin(\tla)~\biass_{\la}.\]
\paragraph{Bounding the variance terms.}
To bound the term $\|\othetnl - \othetl\|_{\Hl(\othetl)}$, we assume $n$ large enough to apply \cref{eq:change-norm-easily} in high probability. Thus, we obtain
\vspace{-5pt}
$$ \|\othetnl - \othetl\|_{\Hl(\othetl)} \leq 2\|\othetnl - \othetl\|_{\Hnl(\othetl)}.
\vspace{-5pt}
$$
Applying \cref{eq:grad_inf_opt} with $\mu = \frac{1}{n} \sum_{i=1}^n \delta_{z_i}$ and $\theta = \othetnl$, since $L_{\mu,\la} = \Lnl$ for the given choice of~$\mu$,
\vspace{-5pt}
$$\|\othetl - \othetnl\|_{\Hnl(\othetl)} ~~\leq~~ \|\nabla \Lnl(\othetl)\|_{\Hnl^{-1}(\othetl)}~/~\dikin( \nm{\othetl-\othetnl}),
\vspace{-5pt}
$$
and applying \cref{eq:change-norm-easily} in high probability again, we obtain
\vspace{-5pt}
$$\|\nabla \Lnl(\othetl)\|_{\Hnl^{-1}(\othetl)} \leq 2\|\nabla \Lnl(\othetl)\|_{\Hl^{-1}(\othetl)}.
\vspace{-5pt}
$$
\paragraph{Bias-variance decomposition.} A technical part of the proof relates $\|\nabla \Lnl(\othetl)\|_{\Hl^{-1}(\othetl)}$ with $\|\nabla \Lnl(\othet)\|_{\Hl^{-1}(\othet)} =: V_\la$, by many applications of Prop.~\ref{prop:self-concordance-main-text}. Here we assume it is done, obtaining
\vspace{-5pt}
$$ \Lo(\othetnl) - \Lo(\othet) ~~\leq~~ \textrm{const.}~\times~(\biass_\la^2 + V_\la^2).
\vspace{-3pt}
$$
\paragraph{From $V_\la$ to $\sqrt{\Deff_\la/n}$.} By construction, $\nabla \Lnl(\othetl) = \frac{1}{n} \sum_{i=1}^n\zeta_i$, with $\zeta_i :=  \nabla\ell_{z_i}(\othetl) + \la \othetl$. Moreover since the $z_i$'s are i.i.d.~samples from $\rho$,  $\Exp{\zeta_i} = \nabla \Ll(\othetl)$. Finally since $\othetl$ is the minimizer of $\Ll$, $\nabla \Ll(\othetl) = 0$. Thus $\nabla \Lnl(\othetl)$ is the average of $n$ i.i.d.~zero-mean random vectors, and so the variance of $V_\la$ is exactly 
\vspace{-5pt}
$$\Exp{V_\la^2} = \frac{1}{n} \Exp{\|\Ho_\la^{-1/2}(\othet)\nabla \ell_Z(\othet) \|^2} = \frac{\Deff_\la}{n}.
\vspace{-5pt}
$$
Finally, by using Bernstein inequality for random vectors \citep[e.g.,][Thm.~3.3.4]{yurinsky1995sums}, we bound $V_\la$ roughly with $\sqrt{\Deff_\la \log(2/\delta)/n}$ in high probability.

\section{Conclusion}
In this paper we have presented non-asymptotic bounds with faster rates than $O(1/\sqrt{n})$, for regularized empirical risk minimization with self-concordant losses such as the logistic loss. It would be interesting to extend our work to algorithms used to minimize the empirical risk, in particular stochastic gradient descent or Newton's method.

\section*{Acknowledgments}
The second author is supported by the ERCIM Alain Bensoussan Fellowship. We acknowledge support from the European Research Council (grant SEQUOIA 724063).

\bibliography{biblio}

\newpage
\appendix

{\Huge{Organization of the Appendix}}

\begin{itemize}
    \item [\textbf{\ref{sec:app-setting}}] \emph{\Large{Setting, definitions, assumptions}}
    \item [\textbf{\ref{sec:gen_sc}}] \emph{\Large{Preliminary results on self concordant losses}}
    \begin{itemize}
        \item [\textbf{\ref{subsec:basicsc}}]\emph{Basic results on self-concordance {\normalfont \bfseries(proof of \cref{prop:self-concordance-main-text})}}
        \item [\textbf{\ref{subsec:loc}}]\emph{Localization properties for $\tla$ {\normalfont \bfseries (proof of \cref{lm:bound-tla-main})}}
    \end{itemize}
    \item [\textbf{\ref{sec:app-main-simple}}] \emph{\Large{Main result, simplified}}
    \begin{itemize}
        \item [\textbf{\ref{sec:main-simple-analytic}}]\emph{Analytic decomposition of the risk}
        \item [\textbf{\ref{sec:main-simple-probabilistic}}]\emph{Concentration lemmas}
        \item [\textbf{\ref{sec:main-simple-final}}]\emph{Final result {\normalfont \bfseries (proof of \cref{thm:gen_bound_222})}}
    \end{itemize}
    \item [\textbf{\ref{sec:app-main-result}}] \emph{\Large{Main result, refined analysis}}
    \begin{itemize}
        \item [\textbf{\ref{sec:analytic_bounds_general}}]\emph{Analytic decomposition of the risk}
        \item [\textbf{\ref{sec:main-analytic-related}}]\emph{Analytic decomposition of terms related to the variance}
        \item [\textbf{\ref{sec:concentration_general}}]\emph{Concentration lemmas}
        \item [\textbf{\ref{sec:final_results_general}}]\emph{Final result {\normalfont \bfseries (proof of \cref{thm:general-result-weak,thm:general-result})}}
    \end{itemize}
    \item [\textbf{\ref{sec:explicit-simple}}] \emph{\Large{Explicit bounds for the simplified case}} {\em\normalfont \bfseries (proof of \cref{cor:base-rates})}
    \item [\textbf{\ref{sec:explicit-general}}] \emph{\Large{Explicit bounds for the refined case}} {\em\normalfont \bfseries(proof of \cref{cor:bounds_source,cor:bounds_source_capacity})}
    \item [\textbf{\ref{sec:additional-lemmas}}] \emph{\Large{Additional lemmas}}
    \begin{itemize}
        \item [\textbf{\ref{sec:lemmas-self}}]\emph{Self-concordance and sufficient conditions to define $L$} 
        \item [\textbf{\ref{sec:lemmas-operators}}]\emph{Bernstein inequalities for operators}
    \end{itemize}
\end{itemize}

\section{Setting, definitions, assumptions}\label{sec:app-setting}

Let $\Z$ be a Polish space and $Z$ a random variable on $\Z$ whith law $\rho$. 
Let $\hh$ be a separable (non-necessarily finite) Hilbert space and let $\ell: \Z \times \hh \to \R$ be a loss function; we denote by $\ell_z(\cdot)$ the function $\ell(z, \cdot)$.
Our goal is to solve
$$ \inf_{\theta \in \hh} \Lo(\theta), \quad \textrm{with} \quad  \Lo(\theta) = \Exp{\ell_Z(\theta)}.$$

Given $(z_i)_{i=1}^n$ we will consider the following estimator
$$\othetnl = \argmin_{\theta \in \hh} \widehat{L}_\la(\theta), \quad \textrm{with} \quad  \Lnl(\theta) := \frac{1}{n}\sum_{i=1}^n \ell_{z_i}(\theta)  + \frac{\la}{2} \|\theta\|^2.$$
The goal of this work is to give upper bounds in high probability to the so called {\em excess risk}
$$ L(\othetnl) - \inf_{\theta \in \hh} L(\theta).$$

In the rest of this introduction we will introduce the basic assumptions required to make $\othetnl$ and the {\em excess risk} well defined, and we will introduce basic objects that are needed for the proofs. 

First we introduce some notation we will use in the rest of the appendix: let $\la \geq 0$, $\theta \in \hh$ and $\mathbf{A}$ be a bounded positive semidefinite Hermitian operator on $\hh$, we denote by $\mathbf{I}$, the identity operator and 
\eqal{\|f\|_{\mathbf{A}} &:= \|\mathbf{A}^{1/2} f\|,\\
\mathbf{A}_\la &:= \mathbf{A} + \la \mathbf{I},\\
\ell_z^\la(\theta) &:= \ell_z(\theta) + \frac{\la}{2}\|\theta\|^2,\\
\Ll(\theta) &:= \Lo(\theta) + \frac{\la}{2}\|\theta\|^2.
}
Now we recall the assumptions we require on the loss function $\ell, \rho$, $(z_i)_{1 \leq i\leq n}$.

\asmiid*
\begin{restatable}[Generalized self-concordance]{ass}{asmgenscb}
\label{asm:gen_scb} The mapping $z \mapsto \ell_z(\theta)$ is measurable for all $\theta \in \hh$ and for any $z \in \Z$, the function $\ell_z$ is convex and three times differentiable. Moreover, there exists a set $\varphi(z) \subset \hh$ such that it holds:
\[\forall \theta \in \hh,~\forall h,k \in \hh,~\left|\nabla^3 \ell_z(\theta)[k,h,h]\right| \leq \sup_{ g \in \varphi(z)}  |k \cdot g|  \ ~\nabla^2 \ell_z(\theta)[h,h]. \]
\end{restatable}
%
%
\asmbounded*
%
%
\asmwd*

Introduce the following definitions.
\bd
Let $\la > 0$, $\theta \in \hh$. We introduce
\eqal{
 \Bone(\theta) &= \sup_{z \in \supp(\rho)} \|\nabla \ell_z(\theta)\|, 
 &\Btwo(\theta) &= \sup_{z \in \supp(\rho)} \Tr\left(\nabla^2 \ell_z(\theta)\right).\\
\Ho(\theta) &= \Exp{\nabla^2 \ell_Z(\theta)}, 
&\Hl(\theta) &= \Ho(\theta) + \la \mathbf{I}.\\
\othetl &= \argmin_{\theta \in \hh} \Ll(\theta).
}
\ed

\bp
Under \cref{asm:gen_scb,asm:bounded,asm:wd}, $\Bone(\theta), \Btwo(\theta), L(\theta), \nabla L(\theta), \Ho(\theta), \othetl$ exist for any $\theta \in \hh, \la >0$. Moreover $\nabla L = \Exp{\nabla \ell_Z(\theta)}$, $\Ho(\theta) =\nabla^2 L(\theta)$ and $\Ho(\theta)$ is trace class.
\ep

\bpr
We start by proving, using the assumptions, that $\Btwo,\Bone$ and $\theta \mapsto \sup_{z \in \supp(\rho)}|\ell_z(\theta)|$ are all locally bounded (see \cref{lm:b2exp,lm:b1exp,lm:b0exp}). This allows us to show that $\ell_z(\theta)$, $\nabla \ell_z(\theta)$ and $\Tr(\nabla^2 \ell_z(\theta))$ are uniformly integrable on any ball of finite radius. The fact that $\othetl$ exists is due to the strong convexity of the function $\Ll$. 
\epr

\bp\label{prop:thetanl-exists}
Under \cref{asm:gen_scb,asm:wd,asm:iid}, when $\la > 0$, $\othetnl$ exists and is unique.
\ep
\bpr
By \cref{asm:iid} we know that $z_1, \dots, z_n$ are in the support of $\rho$. Thus, by \cref{asm:wd}, $\frac{1}{n}\sum_{i=1}^n{\ell_{z_i}}$ is finite valued in $0$. Since $\frac{1}{n}\sum_{i=1}^n{\ell_{z_i}}$ is convex three times differentiable as a sum of such functions, it is real-valued on $\hh$ and hence $\Lnl$ is real-valued on $\hh$;  by strong convexity, $\othetnl$ exists and is unique. 
\epr

Recall that we also make the following regularity assumption.

\asmminimizer*

Finally we conclude with the following definitions that will be used later.

\bd For $\theta \in \hh$, denote by $\nm{\theta}$ the function 
$$\nm{\theta} = \sup_{z \in \supp(\rho)} \left(\sup_{g \in \varphi(z)}|\theta \cdot g|\right),$$
and define
\eqal{
\biass_\la &= \|\nabla \Ll(\othet)\|_{\Hl^{-1}(\othet)},\\
\bvar_\la &=  \|\Hess_\la^{1/2}(\othetl) \Hnl^{-1/2}(\othetl)\|^2~\|\nabla \Lnl(\othetl)\|_{\Hess_\la^{-1}(\othetl)},\\
\Deff_\la &= \Exp{\|\nabla \ell_Z(\othet)\|^2_{\Hl^{-1}(\othet)}},\\
\tla &=  \nm{\othet - \othetl},\\
\rla(\theta) & \quad \textrm{such that}\quad 1/\rla(\theta) = \sup_{z \in \supp(\rho)} \left(\sup_{g \in \varphi(z)} \|g\|_{\Hl^{-1}(\theta)}\right).
}
\ed

\section{Preliminary results on self concordant losses}\label{sec:gen_sc}
In this section, we show how our definition/assumption of self concordance (see \cref{asm:gen_scb}) enables a fine control on the excess risk. In particular, we clearly relate the difference in function values to the quadratic approximation at the optimum as well as the renormalized gradient. We start by presenting a general bounds in \cref{subsec:basicsc} before applying them to the problem of localizing the optimum \cref{subsec:loc}.

\subsection{\label{subsec:basicsc}Basic results on self-concordance }
In this section, as in the rest of the appendix, we are under the conditions of \cref{asm:gen_scb}. \textbf{In this section only}, we give ourselves a probability measure $\mu$ on $\Z$. We will apply the results of this section to $\mu=\rho,\rhohat,\delta_z$, where $\rhohat = \frac{1}{n}\sum_{i=1}^n{\delta_{z_i}}$ and $z$ is sampled from $\rho$.

First of all, let us introduce the following notation. For any probability measure $\mu$ on $\Z$ and any $\theta \in \hh$, define 
\begin{itemize}
    \item $\Rad^{\mu} = \sup_{z \in \supp(\mu)} \left(\sup_{g \in \varphi(z)}\|g\|\right)$,
    \item $\nmm{\theta} = \sup_{z \in \supp(\mu)} \left(\sup_{g \in \varphi(z)}\left|\theta \cdot g\right|\right)$.
\end{itemize}

In order to be able to define $L_\mu(\theta) = \Expb{\mu}{\ell_z(\theta)}$ and to derive under the expectation, we assume that \cref{asm:bounded,asm:wd} are satisfied for $\mu$ (replace $\rho$ by $\mu$ in the assumption).

Since $\mu$ and $\ell$ satisfy \cref{asm:gen_scb,asm:bounded,asm:wd}, \cref{prp:wd} ensures that we can define $L_{\mu}(\theta) = \Expb{\mu}{\ell_z(\theta)}$ and $L_{\mu,\la}(\theta) = L_\mu(\theta) + \frac{\la}{2}\|\theta\|^2$, as well as their respective Hessians $\Hess_{\mu}(\theta)$ and $\Hess_{\mu,\la}(\theta)$. \\\

The following result is greatly inspired from results in \citep{bach2010self} on generalized self concordant losses, and their refinement in \citep{ostrovskii2018finite}. However, while \cref{hess_control,grad_control2,function_values_control} appear more or less explicitly, \cref{grad_control1} provides an easier way to deal with certain bounds afterwards and was not used in this form before.

\bp[using the self-concordance of $\ell$]\label[proposition]{p:cool}
Let $\theta_0,\theta_1 \in \hh$ and $\la \geq 0$. Assume that $(\ell_z)_z$ and $\mu$ satisfy \cref{asm:gen_scb,asm:bounded,asm:wd}. We have the following inequalities:

\begin{itemize}

\item Bounds on Hessians 
\eqal{\label{hess_control}
\Hess_{\mu,\la}(\theta_1)\preceq \exp\left(\nmm{\theta_1-\theta_0}\right)\Hess_{\mu,\la}(\theta_0).
}

\item Bounds on gradients (if $\la > 0$)
\eqal{\label{grad_control1}
&\dikin\left(\nmm{\theta_1-\theta_0}\right) ~\|\theta_1 - \theta_0\|_{\Hess_{\mu,\la}(\theta_0)}\leq \|\nabla L_{\mu,\la}(\theta_1) - \nabla L_{\mu,\la}(\theta_0)\|_{\Hess_{\mu,\la}^{-1}(\theta_0)}, \\
\label{grad_control2}  
& \|\nabla L_{\mu,\la}(\theta_1) - \nabla L_{\mu,\la}(\theta_0)\|_{\Hess_{\mu,\la}^{-1}(\theta_0)}\leq  \dikins\left(\nmm{\theta_1-\theta_0}\right)~\|\theta_1 - \theta_0\|_{\Hess_{\mu,\la}(\theta_0)},
}
where $\dikins(t) = (e^t - 1) / t$ and $\dikin(t) = (1- e^{-t})/t$.
\item Bounds on function values
\eqal{\label{function_values_control}
L_{\mu,\la}(\theta_1) - L_{\mu,\la}(\theta_0) - \nabla L_{\mu,\la}(\theta_0)(\theta_1 - \theta_0) \leq \psi\left( \nmm{\theta_1-\theta_0}\right)\|\theta_1 - \theta_0\|^2_{\Hess_{\mu,\la}(\theta_0)},
}
where $\psi(t) = (e^t - t- 1)/t^2$.
\end{itemize}
\ep

\bpr First of all, note that for any $\mu$ and $\la$, given $\theta \in \hh$ and $k,h \in \hh$,

\eqals{
\left|\nabla^3 L_{\mu,\la}(\theta)[h,k,k]\right|  &= \left|\Expb{z\sim \mu}{\nabla^3 \ell^{\la}_z(\theta)[h,k,k]}\right| \\
&\leq \Expb{z\sim \mu}{\left|\nabla^3 \ell_z(\theta)[h,k,k]\right|} \\
&\leq \Expb{z\sim \mu}{\sup_{g \in \varphi(z)}\left|h \cdot g\right|~\nabla^2 \ell_z(\theta)[k,k]}\\
& \leq \nmm{h}~\Expb{z\sim \mu}{\nabla^2 \ell_z(\theta)[k,k]}  = \nmm{h}~\nabla^2 L_{\mu}(\theta)[k,k] .
}
This yields the following fundamental inequality : 
\eqal{\label{eq:hahaha}\left|\nabla^3 L_{\mu,\la}(\theta)[h,k,k]\right| \leq  ~\nmm{h}~\nabla^2 L_{\mu,\la}(\theta)[k,k] . }
We now define, for any $t \in \R$, $\theta_t := \theta_0 + t(\theta_1 - \theta_0)$.

\paragraph{Point 1.} For the first inequality, let $h \in \hh$ be a fixed vector, and consider the function
    $\varphi : t \in \R \mapsto \nabla^2 L_{\mu,\la}(\theta_t)[h,h]$. Since $\varphi^{\prime}(t) = \nabla^3 L_{\mu,\la}(\theta_t)[\theta_1 - \theta_0,h,h]$, using \cref{eq:hahaha}, we get that $\varphi^{\prime}(t) \leq \nmm{\theta_1-\theta_0}~\varphi(t)$. Using \cref{lm:gronwall}, we directly find that $\varphi(1) \leq \exp(\nmm{\theta_1-\theta_0}) \varphi(0)$, which, rewriting the definition of $\varphi$, yields
    \[\nabla^2 L_{\mu,\la}(\theta_1)[h,h] \leq \exp( \nmm{\theta_1-\theta_0}) \nabla^2 L_{\mu,\la}(\theta_0)[h,h].\]
    This being true for any direction $h$, we have \eqref{hess_control}.
    
\paragraph{ Point 2.} To prove \cref{grad_control1}, let us look at the quantity $(\theta_1 - \theta_0)\cdot \left(\nabla L_{\mu,\la}(\theta_1) - \nabla L_{\mu,\la}(\theta_0)\right) $. Since
    $\nabla L_{\mu,\la}(\theta_1) - \nabla L_{\mu,\la}(\theta_0) = \int_{0}^1{\nabla^2 L_{\mu,\la}(\theta_t) (\theta_1 - \theta_0)dt}$, we have 
    \[(\theta_1 - \theta_0)\cdot \left(\nabla L_{\mu,\la}(\theta_1) - \nabla L_{\mu,\la}(\theta_0)\right) = \int_{0}^1{\nabla^2 L_{\mu,\la}(\theta_t) [\theta_1 - \theta_0,\theta_1 - \theta_0]dt}.\]
    Applying \cref{hess_control} to $\theta_0 $ and $\theta_t$ and the reverse, we find that \[\forall t\in[0,1],~ e^{-t \nmm{\theta_1-\theta_0}}\nabla^2 L_{\mu,\la}(\theta_0) \preceq \nabla^2 L_{\mu,\la}(\theta_t) .  \]
    Hence, integrating the previous equation, we have
    \[(\theta_1 - \theta_0)\cdot\left(\nabla L_{\mu,\la}(\theta_1) - \nabla L_{\mu,\la}(\theta_0)\right) \geq \dikin\left(\nmm{\theta_1-\theta_0}\right) \|\theta_1-\theta_0\|^2_{\Hess_{\mu,\la}(\theta_0)}.\]
    Finally, bounding $(\theta_1 - \theta_0)\cdot\left(\nabla L_{\mu,\la}(\theta_1) - \nabla L_{\mu,\la}(\theta_0)\right)$ by $\|\theta_1 - \theta_0\|_{\Hess_{\mu,\la}(\theta_0)}~\|\nabla L_{\mu,\la}(\theta_1) - \nabla L_{\mu,\la}( \theta_0)\|_{\Hess^{-1}_{\mu,\la}(\theta_0)}$, and simplifying by $\|\theta_1 - \theta_0\|_{\Hess_{\mu,\la}(\theta_0)}$, we obtain \cref{grad_control1}.

\paragraph{Point 3.} To prove \cref{grad_control2}, first write
    \eqals{\|\nabla L_{\mu,\la}(\theta_1) - \nabla L_{\mu,\la}( \theta_0)\|_{\Hess^{-1}_{\mu,\la}(\theta_0)} & = \| \int_{0}^1{\Hess^{-1/2}_{\mu,\la}(\theta_0)\Hess_{\mu,\la}(\theta_t)(\theta_1 - \theta_0)dt}\| & \\
    & = \| \int_{0}^1{\Hess^{-1/2}_{\mu,\la}(\theta_0)\Hess_{\mu,\la}(\theta_t)\Hess^{-1/2}_{\mu,\la}(\theta_0)~\Hess^{1/2}_{\mu,\la}(\theta_0)(\theta_1 - \theta_0)dt}\|
    & \\
    &\leq \left(\int_{0}^1{\|\Hess^{-1/2}_{\mu,\la}(\theta_0)\Hess_{\mu,\la}(\theta_t)\Hess^{-1/2}_{\mu,\la}(\theta_0)\|~dt}\right)~\|\theta_1 - \theta_0\|_{\Hess_{\mu,\la}(\theta_0)}.
    }
    Then apply \cref{hess_control} to have
    \[\forall t\in[0,1],~ \Hess_{\mu,\la}(\theta_t)  \preceq e^{t \nmm{\theta_1-\theta_0}}\Hess_{\mu,\la}(\theta_0) . \]
    This implies 
     \[\forall t\in[0,1],~   \Hess^{-1/2}_{\mu,\la}(\theta_0) \Hess_{\mu,\la}(\theta_t) \Hess^{-1/2}_{\mu,\la}(\theta_0)   \preceq e^{t \nmm{\theta_1-\theta_0}} ~I . \]
     And hence in particular 
     \[\forall t \in [0,1],~ \|\Hess^{-1/2}_{\mu,\la}(\theta_0)\Hess_{\mu,\la}(\theta_t)\Hess^{-1/2}_{\mu,\la}(\theta_0)\| \leq e^{t \nmm{\theta_1-\theta_0}}.\]
     Finally, integrating this, we get
     \[\int_{0}^1{\|\Hess^{-1/2}_{\mu,\la}(\theta_0)\Hess_{\mu,\la}(\theta_t)\Hess^{-1/2}_{\mu,\la}(\theta_0)\|~dt} \leq \dikins\left(\nmm{\theta_1-\theta_0}\right).\]
   Thus \cref{grad_control2} is proved.

\paragraph{Point 4.} To prove \cref{function_values_control}, define $\forall t \in \R,~\varphi(t) = L_{\mu,\la}(\theta_t) - L_{\mu,\theta}(\theta_0) - t~\nabla L_{\mu,\la}(\theta_0)(\theta_1 - \theta_0)$
    We have $\varphi^{\prime \prime}(t) = \|\theta_1-\theta_0\|_{\Hess_{\mu,\la}(\theta_t)}^2 \leq e^{t~\nmm{\theta_1-\theta_0}}\varphi^{\prime \prime}(0)$.
    Then using the fact that $\varphi(0),\varphi^{\prime}(0) = 0$ and integrating this inequality two times, we get the result.
\epr

\bpr {\normalfont \bfseries of \cref{prop:self-concordance-main-text}.}
First note that since the support of $\mu$ is included in the support of $\rho$, \cref{asm:bounded} and \cref{asm:wd} also hold for $\mu$. Hence,  since \cref{asm:bounded,asm:gen_sc,asm:wd} are satisfied, by \cref{prp:wd}, $L_{\mu,\la}$, $\nabla L_{\mu,\la}$ and $\nabla^2 L_{\mu,\la}$ are well-defined. 

Assuming the existence of a minimizer $\othet_{\mu,\la}$ of $L_{\mu,\la}$,
 the reported equations are the same than those of \cref{p:cool} when taking $\theta_1=\theta$ and $\theta_0 = \othet_{\mu,\la}$, with the fact that $\nmm{v} \leq \nm{v}$ for any $v \in \hh$ since the support of $\mu$ is a subset of the support of $\rho$, and $\nabla L_{\mu,\la}(\othet_{\mu,\la}) = 0$. 
 Note that since $L_{\mu,\la}$ is defined on $\hh$, if $\la >0$, then $\othet_{\mu,\la}$ always exists and is unique by strong convexity.
\epr

\subsection{Localization properties for $\tla$}\label{subsec:loc}

The aim of this section is to localize the optima $\othetl$ and $\othetnl$ using the re-normalized gradient. This type of result is inspired by Proposition 2 of \citep{bach2010self} or Proposition 3.5 of \citep{ostrovskii2018finite}. However, their proof is based on a slightly different result, namely \cref{grad_control1}, and its formulation is slightly different. Indeed, while the two propositions mentioned above concentrate on performing a quadratic approximation directly, we bound the term that could have been too large in that quadratic approximation.

\bp[localisation]\label[proposition]{prop:localization} Let $\theta \in \hh$, then the following holds
\eqal{\label{cond:loc_exp}
\|\nabla \Ll(\theta)\|_{\Hess_\la^{-1}(\theta)} \leq \frac{\rla(\theta)}{2 } \quad &\implies \quad \nm{\theta - \othetl } = \tla \leq \log 2,\\
\label{cond:loc_emp} \|\nabla \Lnl(\theta)\|_{\Hl^{-1}(\theta)}~\|\Hnl^{-1/2}(\theta) \Hl^{1/2}(\theta)\|^2 \leq \frac{\rla(\theta)}{2} \quad &\implies \nm{\theta - \othetnl } \leq \log 2. 
}
\ep

\bpr
To prove \cref{cond:loc_exp}, we first write 
\[\nm{\theta - \othetl } =  \sup_{z \in \supp(\rho)} \sup_{g \in \varphi(z)}\left|(\theta - \othetl)\cdot g\right| \leq  \|\theta - \othetl \|_{\Hl(\theta)}~\sup_{z \in \supp(\rho)} \sup_{g \in \varphi(z)}\|g\|_{\Hess_\la^{-1}(\theta)}.\]
Now we use \cref{eq:grad_inf_opt} to bound $\|\theta - \othetl \|_{\Hl(\theta)}$, and putting things together, we get
\[\nm{\theta - \othetl } \dikin\left(\nm{\theta-\othetl}\right) \leq  \frac{\|\nabla \Ll(\theta)\|_{\Hess_\la^{-1}(\theta)}}{\rla(\theta)}.\]
Using the fact that $t\dikin(t) = 1 - e^{-t}$ is an increasing function, we see that if $t \dikin(t) \leq 1/2$, then $t \leq \log 2$ hence the result.\\\

To prove \cref{cond:loc_emp}, we use the same reasoning. First, we bound 
\[\nm{\theta-\othetnl} =  \sup_{z \in \supp(\rho)}\sup_{g \in \varphi(z)}\left|(\theta - \othetnl)\cdot g\right| \leq  \|\theta - \othetnl \|_{\Hnl(\theta)}~\|\Hnl^{-1/2}(\theta) \Hl^{1/2}(\theta)\|~\sup_{z \in \supp(\rho)}\sup_{g \in \varphi(z)}\|g\|_{\Hess_\la^{-1}(\theta)}.\]
Now using \cref{eq:grad_inf_opt} to the function $\Lnl$, we get 
\[\nm{\theta-\othetnl} \dikin\left(\mathsf{t}^{\hat{\rho}}(\theta - \othetnl)\right) \leq  \|\nabla \Lnl(\theta) \|_{\Hnl^{-1}(\theta)}~\|\Hnl^{-1/2}(\theta) \Hl^{1/2}(\theta)\|~\frac{1}{\rla(\theta)}. \]
Now using the fact that $\mathsf{t}^{\hat{\rho}}(\theta - \othetnl) \leq \nm{\theta-\othetnl}$ and that $\dikin$ is a decreasing function, and that $\|\nabla \Lnl(\theta) \|_{\Hnl^{-1}(\theta)} \leq \|\Hnl^{-1/2}(\theta) \Hl^{1/2}(\theta)\|~\|\nabla \Lnl(\theta) \|_{\Hl^{-1}(\theta)}$, this yields
\[\nm{\theta-\othetnl} \dikin\left(\nm{\theta-\othetnl}\right) \leq \|\nabla \Lnl(\theta) \|_{\Hl^{-1}(\theta)}~\|\Hnl^{-1/2}(\theta) \Hl^{1/2}(\theta)\|^2~\frac{1}{\rla(\theta)}. \]
We conclude using the same argument as before. 
\epr

\section{Main result, simplified}\label{sec:app-main-simple}

In this section, we perform a simplified analysis in the case where we assume nothing on $\biass_\la$ more than just the fact that $\othet$ exists. 
In this section we assume that $\ell_z$ and $\rho$ satisfy \cref{asm:gen_scb,asm:bounded,asm:wd,asm:minimizer}. 

\bd[Definition of $\Bob$, $\Btb$ and $\Deffsup_\la$]\label[definition]{asm:strong_wd}
Under assumptions \cref{asm:gen_scb,asm:bounded,asm:wd,asm:minimizer}, the following quantities are well-defined and real-valued.
\[\Bob= \sup_{\|\theta\| \leq \|\othet\|} \Bone(\theta) \,\quad \Btb = \sup_{\|\theta\|\leq \|\othet\|}\Btwo(\theta), \quad \Deffsup_\la = \Exp{\|\nabla \ell_z(\othetl)\|^2_{\Hl^{-1}(\othetl)}}.\]
\ed

\bp
The quantities in \cref{asm:strong_wd} are finite and moreover 
$$\Deffsup_\la \leq \frac{\Bob^2}{\la}.$$
\ep
\bpr
These are well defined thanks to \cref{lm:b2exp,lm:b1exp}.
\epr

\bd[Constants]\label[definition]{rmk:constants2}
In this section, we will use the following constants.
$$\Kv= \frac{1 + \psi(\log 2)}{\dikin(\log 2)^2} \leq 4, \quad \triangle = 2\sqrt{2}\left(1 + \frac{1}{2 \sqrt{3}}\right) \leq 4, $$
$$ \Cb = 1 + \frac{\Kv}{8}  \leq 2 ,\qquad \Cv =2  \Kv \triangle^2 \leq 84. $$
\ed

\subsection{Analytic results}\label{sec:main-simple-analytic}

\bt[Analytic decomposition] \label[theorem]{prp:anal_dec_simple}For any $\la >0$ and $n \in \N$, if $\frac{\Rad}{\sqrt{\la}} \bvar_\la \leq \frac{1}{2}$, 
\eqal{\label{eq:an_dec_2}
L(\othetnl) - L(\othet) \leq \Kv~\bvar_\la^2 + \lambda \|\othet\|^2,}
where $\Kv$ is defined in \cref{rmk:constants2}.
\et

\bpr

First decompose the excess risk of $\othetnl$ in the following way:
$$ \Lo(\othetnl) - \Lo(\othet) = \underbrace{\Ll(\othetnl) - \Ll(\othetl)}_{\text{variance}} + \underbrace{\Lo(\othetl) - \Lo(\othet)}_{\text{bias}} + \underbrace{\frac{\la}{2}\left(\|\othetl\|^2 - \|\othetnl\|^2\right) }_{\text{mixed}}.$$

\noindent{\normalfont \bfseries 1) Variance term:} For the variance term, use \cref{eq:bound_to_opt}
    \[\Ll(\othetnl) - \Ll(\othetl) \leq \psi\left(\nm{\othetl - \othetnl}\right)\|\othetnl - \othetl\|_{\Hl(\othetl)}^2.\]
\noindent{\normalfont \bfseries 2) Bias term:} For the bias term, note that since $\|\othetl\| \leq \|\othet\|$,
    \[L(\othetl) - L(\othet) = \Ll(\othetl) - \Ll(\othet) + \frac{\la}{2}\|\othet\|^2 - \frac{\la}{2}\|\othetl\|^2 \leq \frac{\la}{2}\|\othet\|^2 . \]
\noindent{\normalfont \bfseries 3) Mixed term:} For the mixed term, since $\|\othetl\|_{\Hl(\othetl)^{-1}} \leq \|\Hl(\othetl)^{-1/2}\|\|\othetl\| \leq \la^{-1/2}\|\othetl\| \leq \la^{-1/2}\|\othet\|$, we have 
    \eqals{\frac{\la}{2}\left(\|\othetl\|^2 - \|\othetnl\|^2\right) & = \frac{\la}{2}\left(\othetl - \othetnl\right)\cdot \left(\othetl + \othetnl\right)\\
    & \leq \frac{\la}{2}\|\othetl - \othetnl\|_{\Hl(\othetl)}~\left(\|\othetnl - \othetl\|_{\Hl(\othetl)^{-1}} + 2\| \othetl\|_{\Hl(\othetl)^{-1}}\right)\\
    &\leq \frac{1}{2}\|\othetl-\othetnl\|^2_{\Hl(\othetl)} + \sqrt{\la}\|\othet\|~\|\othetl-\othetnl\|_{\Hl(\othetl)}\\
    &\leq \|\othetl-\othetnl\|^2_{\Hl(\othetl)} + \frac{\la}{2}\|\othet\|^2.
    }
where we get the last inequality by using $ab \leq \frac{a^2}{2} + \frac{b^2}{2}$.

\noindent{\normalfont \bfseries 4) Putting things together}

\[L(\othetnl) - L(\othet) \leq \left(1+\psi\left(\nm{\othetl - \othetnl}\right)\right)\|\othetl-\othetnl\|^2_{\Hl(\othetl)} + \lambda \|\othet\|^2.\]
By using \cref{eq:grad_inf_opt} we have
\eqals{
\|\othetl-\othetnl\|_{\Hl(\othetl)} & \leq \|\Hl^{1/2}(\othetl)\Hnl^{-1/2}(\othetl)\|~\|\othetl-\othetnl\|_{\Hnl(\othetl)}  \\
&\leq \frac{1}{\dikin\left(\mathsf{t}^{\widehat{\rho}}(\othetl - \othetnl)\right)}~\|\Hl^{1/2}(\othetl)\Hnl^{-1/2}(\othetl)\|~\|\nabla \Lnl(\othetl)\|_{\Hnl^{-1}(\othetl)}.
}
Note that by multiplying and dividing for $\Hl^{1/2}(\othetl)$, 
\eqals{
\|\nabla \Lnl(\othetl)\|_{\Hnl^{-1}(\othetl)} &= \|\Hnl^{-1/2}(\othetl)\nabla \Lnl(\othetl)\| = \|\Hnl^{-1/2}(\othetl)\Hl^{-1/2}(\othetl) \Hl^{1/2}(\othetl)\nabla \Lnl(\othetl)\| \\
&\leq \|\Hnl^{-1/2}(\othetl)\Hl^{-1/2}(\othetl)\| \|\Hl^{1/2}(\othetl)\nabla \Lnl(\othetl)\| \\
& = \|\Hnl^{-1/2}(\othetl)\Hl^{-1/2}(\othetl)\| \|\nabla \Lnl(\othetl)\|_{\Hl^{-1}(\othetl)}.
}
Then,
\eqals{
\|\othetl-\othetnl\|_{\Hl(\othetl)} &\leq \frac{1}{\dikin\left(\nm{\othetl - \othetnl}\right)}~\|\Hl^{1/2}(\othetl)\Hnl^{-1/2}(\othetl)\|^2~\|\nabla \Lnl(\othetl)\|_{\Hl^{-1}(\othetl)} \\
&=\frac{1}{\dikin\left(\nm{\othetl - \othetnl}\right)}~\bvar_\la.
}
Now we know that using \cref{cond:loc_emp}, if $\bvar_\la \leq \frac{\rla(\othetl)}{2}$, then $\nm{\othetl - \othetnl} \leq \log 2$, which yields the following bound:
\[L(\othetnl) - L(\othet) \leq \frac{\left(1+\psi\left(\log 2\right)\right)}{\dikin(\log 2)^2}\bvar_\la + \lambda \|\othet\|^2.\]
Finally, we can bound $\frac{1}{\rla(\othetl)}\leq \frac{\Rad}{\la^{1/2}}$ to have the final form of the proposition.

\epr

\subsection{Probabilistic results} \label{sec:main-simple-probabilistic}

\blm[bounding $\|\nabla \Lnl(\othetl)\|_{\Hl(\othetl)}$]\label[lemma]{lm:bndind} Let $n \in \N$, $\la >0$ and $\delta \in (0,1]$. For $k \geq 1$, if
\[n \geq 24 \frac{\Btb}{\la} \log \frac{2}{\delta}, n \geq k^2 2\log \frac{2}{\delta},\]
then with probability at least $1-\delta$,
 \[\|\Hl(\othetl)^{-1/2}\nabla \Lnl(\othetl)\|  \leq\triangle /2 \sqrt{\frac{  \Deffsup_\la\vee (\Bob^2/\Btb)~\log \frac{2}{\delta}}{  n}} + \frac{2}{k} \sqrt{\la}\|\othet\|\]
 where $\triangle$ is defined in \cref{rmk:constants2}.
\elm
\bpr
\noindent{\normalfont \bfseries 1) }First use Bernstein inequality for random vectors \citep[e.g. Thm.~3.3.4 of][]{yurinsky1995sums}: for any $n \in \N$ and $\delta \in (0,1]$, with probability at least $1 - \delta$, we have 
    $$ \|\Hl(\othetl)^{-1/2}\nabla \Lnl(\othetl)\| \leq \frac{2 M \log \frac{2}{\delta}}{n} + \sigma~\sqrt{\frac{2 \log \frac{2}{\delta}}{n}},$$
    where $M = \sup_{z \in \supp(\rho)} \|\nabla \ell_z^{\la}(\othetl)\|_{\Hl^{-1}(\othetl)}$ and $\sigma = \Exp{\|\nabla \ell_z^{\la}(\othetl)\|_{\Hl^{-1}(\othetl)}^2}^{1/2}$.\\\
    
    \noindent{\normalfont \bfseries 2) }Using the fact that $\nabla \ell_z^{\la}(\othetl) = \nabla \ell_z(\othetl) + \la \othetl$, we bound $M$ as follows:
    \[ M = \sup_{z \in \supp(\rho)}{\|\nabla \ell_z^{\la}(\othetl)\|_{\Hl(\othetl)}} \leq  \sup_{z \in \supp(\rho)}{\|\nabla \ell_z(\othetl)\|_{\Hl(\othetl)}} + \la \|\othetl\|_{\Hl^{-1}(\othetl)} \leq  \frac{\Bob}{\sqrt{\la}} + \sqrt{\la}\|\othet\|,   \]
    where in the last inequality, we use the fact that $\|\othetl\|_{\Hl^{-1}(\othetl)} \leq \frac{1}{\sqrt{\la}}\|\othetl\| \leq \frac{1}{\sqrt{\la}}\|\othet\|$. Similarly, we bound $\sigma$
    \[\sigma \leq \Exp{\|\nabla \ell_z(\othetl)\|_{\Hl^{-1}(\othetl)}^2}^{1/2} + \la \|\othetl\|_{\Hl^{-1}(\othetl)} \leq \Deffsup_\la^{1/2} + \sqrt{\la}\|\othet\|.\]
    \noindent{\normalfont \bfseries 3) }Injecting these bounds in the concentration inequality, 
    \eqals{
    \|\Hl(\othetl)^{-1/2}\nabla \Lnl(\othetl)\| &\leq \sqrt{ \frac{2 \Btb ~\log \frac{2}{\delta}}{\la n}}\sqrt{\frac{2(\Bob^2 / \Btb)~\log \frac{2}{\delta}}{ n}} + \sqrt{\frac{2 \Deffsup_\la~\log \frac{2}{\delta}}{ n}}\\
    &+ \sqrt{\la}\|\othet\|\left(\frac{2 \log \frac{2}{\delta}}{n}+\sqrt{\frac{2 \log \frac{2}{\delta}}{n}}\right),}
    where we have decomposed $\frac{2 \Bob^2 \log \frac{2}{\delta}}{\sqrt{\la}n} = \sqrt{ \frac{2 \Btb ~\log \frac{2}{\delta}}{\la n}}\sqrt{\frac{2(\Bob^2 / \Btb)~\log \frac{2}{\delta}}{ n}}$ for the first term. Reordering the terms, this yields
    \eqals{
    \|\Hl(\othetl)^{-1/2}\nabla \Lnl(\othetl)\| & \leq \left(1 +\sqrt{ \frac{2 \Btb ~\log \frac{2}{\delta}}{\la n}}\right)\sqrt{\frac{2  \Deffsup_\la \vee\left(\Bob^2/\Btb\right)~\log \frac{2}{\delta}}{ n}}\\
     &+ \sqrt{\la}\|\othet\|\left(\frac{2 \log \frac{2}{\delta}}{n}+\sqrt{\frac{2 \log \frac{2}{\delta}}{n}}\right).
    }
    
\noindent{\normalfont \bfseries 4) } Now assuming that     
\[n \geq 24 \frac{\Btb}{\la} \log \frac{2}{\delta}, n \geq k^2 2 \log \frac{2}{\delta},\]
this yields
 \eqals{
    \|\Hl(\othetl)^{-1/2}\nabla \Lnl(\othetl)\|  \leq \left(1 + \frac{1}{2 \sqrt{3}}\right)\sqrt{\frac{2  \Deffsup_\la \vee (\Bob^2/\Btb)~\log \frac{2}{\delta}}{  n}} + \frac{2}{k}\sqrt{\la}\|\othet\|.
    }
\epr

Combining the two previous lemmas, we get:

\blm[Bounding $\bvar_\la$]\label[lemma]{lm:bound_b_var} Let $n \in \N$ and $0 < \la \leq \Btb $. Let $\delta \in (0,1]$. If for $k\geq 1$

\[n\geq 24 \frac{\Btb}{\la} \log \frac{8 \Btb}{\la \delta},~~~ n \geq 2 k^2 \log \frac{2}{\delta},\]
then with probability at least $1 - 2 \delta$, 
\[\bvar_\la \leq \triangle ~\sqrt{\frac{\Deffsup_\la \vee (\Bob^2/\Btb)~\log \frac{2}{\delta}}{  n}} + \frac{4}{k} \sqrt{\la}\|\othet\|,\]
where $\triangle$ is a constant defined in \cref{rmk:constants2}.
\elm

\bpr
Recall that  $\bvar_\la =\|\Hl^{1/2}(\othetl)\Hnl^{-1/2}(\othetl)\|^2~\|\nabla \Lnl(\othetl)\|_{\Hnl^{-1}(\othetl)}$. Using \cref{lm:hessian_concentration}, under the conditions of this lemma, we have $\|\Hl^{1/2}(\othetl)\Hnl^{-1/2}(\othetl)\|^2 \leq 2$. Combining this with the bound for $\|\nabla \Lnl(\othetl)\|_{\Hnl^{-1}(\othetl)}$ obtained in \cref{lm:bndind}, we get the result (the probability $1-2\delta$ comes from the fact that we perform a union bound). 

\epr

\subsection{Final result}\label{sec:main-simple-final}

\bt[General bound, simplified setting]\label[theorem]{thm:gen_bound_22}
Let $n \in \N$ and $0 < \la \leq \Btb$. Let $\delta \in (0,1]$. If 

\[ n \geq 512 \left( \|\othet\|^2 \Rad^2\vee 1\right) \log \frac{2}{\delta},~~~n\geq 24 \frac{\Btb}{\la} \log \frac{8 \Btb}{\la \delta}, ~~~ n \geq 16 \triangle^2 \Rad^2 \frac{\Deffsup_\la \vee (\Bob^2/\Btb)}{ \la } ~\log \frac{2}{\delta}, \]
then with probability at least $1 - 2 \delta$, 
\eqals{
L(\othetnl) - L(\othet) \leq \Cv~ \frac{\Deffsup_\la \vee (\Bob^2 /\Btb)}{ n}~\log \frac{2}{\delta} +\Cb \lambda \|\othet\|^2,}
where $\triangle,\Cb,\Cv$ are defined in \cref{rmk:constants2}.
\et

\bpr 
\noindent{\normalfont \bfseries 1)}  Recall the analytical decomposition in \cref{prp:anal_dec_simple}. For any $\la >0$ and $n \in \N$, if $\frac{\Rad}{\sqrt{\la}} \bvar_\la \leq \frac{1}{2}$, 
\eqals{
L(\othetnl) - L(\othet) \leq \Kv~\bvar_\la^2 + \lambda \|\othet\|^2,}
where $\Kv$ is defined in \cref{rmk:constants2}.

\noindent{\normalfont \bfseries 2)}  Now apply \cref{lm:bound_b_var} for a given $k\geq 1$. If

\[n \geq 24 \frac{\Btb}{\la}\log \frac{8 \Btb}{\la \delta},~~~ n \geq 2 k^2 \log \frac{2}{\delta},\]
then with probability at least $1 - 2 \delta$, 
\[\bvar_\la \leq \triangle ~\sqrt{\frac{\Deffsup_\la \vee (\Bob^2/\Btb)~\log \frac{2}{\delta}}{ n}} + \frac{4}{k} \sqrt{\la}\|\othet\|,\]
where $\triangle$ is a constant defined in \cref{rmk:constants2}.

In order to satisfy the condition to have the analytical decomposition, namely $\frac{\Rad}{\sqrt{\la}} \bvar_\la \leq \frac{1}{2}$, it is therefore sufficient to have 
\[\triangle ~\Rad~\sqrt{\frac{\Deffsup_\la \vee (\Bob^2/\Btb)~\log \frac{2}{\delta}}{\la  n}} \leq \frac{1}{4},\qquad \frac{4}{k}  \Rad\|\othet\| \leq \frac{1}{4} .\]

\noindent{\normalfont \bfseries 3)} Thus, if we choose $k = 16 (\Rad \|\othet\| \vee 1)$, we have both $k \geq 1$ and the second condition in the previous equation. Moreover, the condition $n \geq 2 k^2 \log \frac{2}{\delta}$ becomes $n \geq 512 (\Rad^2 \|\othet\|^2 \vee 1) \log \frac{2}{\delta}$. Hence, under the conditions of this theorem, we can apply the analytical decomposition :
\eqals{
L(\othetnl) - L(\othet) \leq \Kv~\bvar_\la^2 + \lambda \|\othet\|^2 \leq 2\Kv \triangle^2 ~\frac{\Deffsup_\la \vee (\Bob^2/\Btb)~\log \frac{2}{\delta}}{ n} + \left(1 + \Kv \frac{32}{k^2}\right) \la \|\othet\|^2.}
In the last inequality, we have used $(a+b)^2 \leq 2a^2 + 2b^2$ to separate the terms coming from $\bvar_\la^2$.

Finally, using the fact that $k \geq 16$ and hence that $\frac{32}{k^2} \leq \frac{1}{8}$, we get the constants in the theorem. 
\epr

\begin{proof} 
\noindent{\normalfont \bfseries of \cref{thm:gen_bound_222}}
Since $\forall \la >0,~\Deff_\la \leq \frac{\Bob^2}{\la}$, and since $\la \leq \Btb$, $\Deff_\la \vee \Bob^2/\Btb \leq \frac{\Bob^2}{\la}$. From \cref{rmk:constants2}, we get that $\triangle \leq 4, ~\Cb \leq 2, ~\Cv \leq 84$. Thus, we can use these bounds in \cref{thm:gen_bound_22} to obtain the result.

\end{proof}

\section{Main result, refined analysis}\label{sec:app-main-result}

In subsection \cref{sec:analytic_bounds_general} we split the excess risk in terms of bias and variance, that will be controlled in \cref{sec:concentration_general}, the final result is \cref{thm:general-result} in \cref{sec:final_results_general}, while in \cref{sec:explicit-general} a version with explicit dependence in $\la, n$ is reported.

\paragraph{Constants}

First, we introduce three constants that will be crucial for the final bound.

\bd
$$\Bos = \Bone(\othet), \qquad \Bts = \Btwo(\othet), \qquad \Qc^* = \Bos/\sqrt{\Bts}.$$
\ed

In the following sections, we also will use the following functions of $\tla$ and $\tlab$ which we will treat as constants (see \cref{prp:constantsmain}).

\bd\label[definition]{df:constantsmain} 
{\small
\eqals{
K_{\textup{bias}}(\tla) &= 2\frac{\psi(\tla + \log 2)}{\dikin(\tla)^2} \leq 2e^{3 \tla} ,
&K_{\textup{var}}(\tla) &= 2 \frac{\psi(\tla + \log 2)e^{\tla}}{\dikin(\log 2)^2} \leq 8e^{2 \tla},\\
\square_1(\tla) &= e^{\tla/2},
&\square_2(\tla) &= e^{\tla/2}\left(1 + e^{\tla}\right) \leq 2 e^{3 \tla/2}\\
\mathsf{C}_{\textup{bias}} &= \psi(\tla+\log 2)\left(\frac{2}{\dikin(\tla)} + \frac{e^{\tla}}{\dikin(\log 2)^2}\right) \leq 6 e^{2 \tla},
&\vspace{-1cm}\mathsf{C}_{\textup{var}} &= \frac{64\psi(\tla+\log 2)e^{2 \tla}}{\dikin(\log 2)^2} \leq 256 e^{3 \tla}\\
\triangle_1 &= 576 \square_1^2 \square_2^2 (1/2 \vee \tlab)^2 \leq 2304 e^{4 \tla}(\tlab \vee 1/2)^2,&\triangle_2 &= 256 \square_1^4 \leq 256 e^{2 \tla}.}
}
\ed

Note that theses functions are all increasing in $\tla$ and $\tlab$, and are lower bounded by strictly positive constants.

For the second bounds, we use the fact that $\psi(t) \leq \frac{e^t}{2}$ and $1/\dikin(t) \leq  e^t$ to bound all the quantities using only exponentials of $\tla$.\\

A priori, these constants will depend on $\la$. However, we can always bound $\tla$ and $\tlab$ in the following way.

\blm\label[lemma]{lm:tla}
Recall the definitions of $\tla := \nm{\othetl-\othet}$ and $\tlab := \frac{\biass_\la}{\rla(\othet)}$. We have the following cases.
\begin{itemize}
    \item If $\tlab \leq \frac{1}{2}$, then $\tla \leq \log 2$,
    \item else, $\tlab \leq \Rad \|\othet\|$ and   $\tla \leq 2 \Rad \|\othet \| $.
\end{itemize}

\elm

\bpr
The first point is a direct application of \cref{cond:loc_exp}. One can obtain the second by noting that $\nm{\othetl - \othet} \leq \Rad \| \othetl - \othet \| $. Since $\|\othetl\| \leq \|\othet\|$, we have the bound on $\tla$. For the bound on $\tlab$, since $\biass_\la \leq \sqrt{\la}\|\othet\|$ and $\frac{1}{\rla(\othet)} \leq \frac{\Rad}{\sqrt{\la}}$, we have the wanted bound.

\epr

Hence, we can always bound the constants in \cref{df:constantsmain} by constants independant of $\la$.

\bp \label[proposition]{prp:constantsmain}
If $\tlab \leq 1/2$, then $\tla \leq \log 2$ and
\eqals{
K_{\textup{bias}}(\tla) &\leq  4, & K_{\textup{var}}(\tla) &\leq 7 , &
\square_1(\tla) &\leq 2, 
&\square_2(\tla) &\leq 5\\
\triangle_1(\tla,\tlab) &\leq 5184, 
&\triangle_2(\tla) &\leq  1024, &
\mathsf{C}_{\textup{bias}} &\leq 6, 
&\mathsf{C}_{\textup{var}} &\leq  414.
}
Else,
\eqals{
K_{\textup{bias}}(\tla) &\leq  2 e^{6 \Rad \| \othet \|}, & K_{\textup{var}}(\tla) &\leq 8 e^{4 \Rad \|\othet \|} , &
\square_1(\tla) &\leq e^{\Rad \|\othet\|},\\ 
\square_2(\tla) &\leq 2 e^{3 \Rad \|\othet\|},
& \triangle_1(\tla,\tlab) &\leq 2304 (\Rad \|\othet\|)^2 e^{8 \Rad \|\othet\|}, 
&\triangle_2(\tla) &\leq  256 e^{4 \Rad \|\othet\|}, \\
\mathsf{C}_{\textup{bias}} &\leq 6e^{4\Rad \|\othet\|}, 
&\mathsf{C}_{\textup{var}} &\leq  256 e^{6\Rad \|\othet\|}.
}
\ep

\bpr
For the first bound, we use the fact that $\tla \leq \log 2$ and plug that in the expressions above as these functions are increasing in $\tla$. We compute them numerically from the definition.\\\

For the second set of bounds, we simply inject the bounds for $\tla$ and $\tlab$ in the second bounds of \cref{df:constantsmain}.
\epr

\subsection{Analytic decomposition of the risk \label{sec:analytic_bounds_general}}

In this section, we make use of self-concordance to control certain quantities required to control the variance, with respect to our main quantities $\biass_\la$, $\rla$ and $\Deff_\la$. The excess risk has been already decomposed in \cref{sec:proof}. \\

\bt[Analytic decomposition]\label[theorem]{prp:anal_dec}
Let $\la > 0$ and $K_{\textup{bias}}$ and $K_{\textup{var}}$ be the \textit{increasing functions} of $\tla$ described in \cref{eq:Kbias-Kvar}. When $\bvar_\la \leq \rla(\othetl)/2$, then 
\eqal{\label{anal_dec}
L(\othetnl) - L(\othet) ~~~\leq~~~  K_{\textup{bias}}(\tla)~\biass_\la^2 ~~+~~ K_{\textup{var}}(\tla) ~\bvar_\la^2.
}
Moreover $K_{\textup{bias}}(\tla), K_{\textup{var}}(\tla) \leq 7$  if  $\biass_\la \leq \frac{1}{2} \rla(\othet)$, otherwise $K_{\textup{bias}}(\tla), K_{\textup{var}}(\tla) \leq 8 e^{6\|\othet\|~\Rad}$ (see \cref{prp:constantsmain} in \cref{sec:app-main-result} for more precise bounds). 
\et
\bpr 
Since $\othet$ exists by \cref{asm:minimizer}, using \cref{eq:bound_to_opt}, applied with $\mu = \rho$ and $\la = 0$, we have $\Lo(\theta) - \Lo(\othet) \leq \psi(\nm{\theta-\othet})\|\theta - \othet\|^2_{\Ho(\othet)}$, for any $\theta \in \hh$. By setting $\theta = \othetnl$, we obtain
\[\Lo(\othetnl) - \Lo(\othet) \leq \psi( \nm{\othetnl-\othet})\|\othetnl - \othet\|^2_{\Ho(\othet)}.\]
Using the fact that $\Ho(\othet)\preceq \Ho(\othet) + \la I =: \Hl(\othet)$, by adding and subtracting $\othetl$, we have
$$\|\othetl - \othet\|_{\Ho(\othet)} \leq \|\othetl - \othet\|_{\Hl(\othet)} \leq \|\othetl - \othet\|_{\Hl(\othet)} + \|\othetnl - \othetl\|_{\Hl(\othet)},$$
and analogously since $\nm{\cdot}$ is a (semi)norm, $\nm{\othetnl-\othet} \leq \tla + \nm{\othetnl-\othet}$, so
$$
\Lo(\othetnl) - \Lo(\othet) ~~\leq~~ \psi(\tla + \nm{\othetnl-\othetl})~(\|\othetl - \othet\|_{\Hl(\othet)} + \|\othetnl - \othetl\|_{\Hl(\othet)})^2.
$$
By applying \cref{eq:hess_control_opt} with $\mu = \rho$ and $\theta = \othet$,  we have $\Hl(\othet) \preceq e^{\tla} \Hl(\othetl)$ and so
\eqal{\label{eq:decomposition}
\begin{aligned}
\Lo(\othetnl) - \Lo(\othet) ~~\leq~~ \psi({\tla + \nm{\othetnl-\othetl}})~(\|\othetl - \othet\|_{\Hl(\othet)} + e^{\tla/2}\|\othetnl - \othetl\|_{\Hl(\othetl)})^2.
\end{aligned}
}
The terms $\tla $ and $\|\othetl - \othet\|_{\Hl(\othet)}$ are related to the \textit{bias terms}, while the terms $ \nm{\othetnl-\othetl}$ and $\|\othetnl - \othetl\|_{\Hl(\othetl)}$ are related to the \textit{variance term}.

\paragraph{Bounding the bias terms.}
Recall the definition of the bias $\biass_\la = \|\nabla\Ll(\othet)\|_{\Hess^{-1}_{\la}(\othet)}$.
We bound $\tla = \nm{\othetl - \othet}$, by \cref{lm:bound-tla-main} and the term $\|\othet - \othetl\|_{\Hl(\othet)}$ by applying \cref{eq:grad_inf_opt} with $\mu = \rho$ and $\theta = \othet$
\[\|\othet - \othetl\|_{\Hl(\othet)}~~\leq~~ 1/\dikin(\tla)~ \|\nabla\Ll(\othet)\|_{\Hess^{-1}_{\la}(\othet)} ~~=~~ 1/\dikin(\tla)~\biass_{\la}.\]
\paragraph{ Bounding the variance terms.}
First we bound the term $\|\othetnl - \othetl\|_{\Hl(\othetl)} := \|\Hl(\othetl)^{1/2} (\othetnl - \othetl)\|$, by multiplying and dividing for $\Hnl(\othetl)^{-1/2}$, we have
\eqals{
\|\othetnl - \othetl\|_{\Hl(\othetl)} &= \|\Hl(\othetl)^{1/2}\Hnl(\othetl)^{-1/2}\Hnl(\othetl)^{1/2} (\othetnl - \othetl)\| \\
&\leq \|\Hl(\othetl)^{1/2}\Hnl(\othetl)^{-1/2}\|\|\othetnl - \othetl\|_{\Hnl(\othetl)}.
}
Applying \cref{eq:grad_inf_opt} with $\mu = \frac{1}{n} \sum_{i=1}^n \delta_{z_i}$ and $\theta = \othetnl$, since $L_{\mu,\la} = \Lnl$ for the given choice of $\mu$, we have
\[\|\othetl - \othetnl\|_{\Hnl(\othetl)} ~~\leq~~ \|\nabla \Lnl(\othetl)\|_{\Hnl^{-1}(\othetl)}~/~\dikin( \nm{\othetl-\othetnl})\]
and since $\|\nabla \Lnl(\othetl)\|_{\Hnl^{-1}(\othetl)} := \|\Hnl^{-1/2}(\othetl)\nabla \Lnl(\othetl)\|$, by multiplying and dividing by $\Hl(\othetl)$, we have:
\eqals{
\|\Hnl^{-1/2}(\othetl)\nabla \Lnl(\othetl)\| &= \|\Hnl^{-1/2}(\othetl)\Hl(\othetl)^{1/2}\Hl(\othetl)^{-1/2}\nabla \Lnl(\othetl)\|  \\
&\leq \|\Hnl^{-1/2}(\othetl)\Hl(\othetl)^{1/2}\|\|\nabla \Lnl(\othetl)\|_{\Hl^{-1}(\othetl)}.
}
Then
\eqals{
\|\othetl - \othetnl\|_{\Hl(\othetl)} ~\leq~ \frac{1}{\dikin( \nm{\othetl-\othetnl})}~\|\Hess_\la^{1/2}(\othetl) \Hl^{-1/2}(\othetl)\|^2~\|\nabla \Lnl(\othetl)\|_{\Hnl^{-1}(\othetl)} = \frac{\bvar_\la}{\dikin( \nm{\othetl-\othetnl})}. }
To conclude this part of the proof we need to bound $\nm{\othetnl-\othetl}$. Since we require $\bvar_\la/\rla(\othetl) \leq 1/2$, by \cref{prop:localization} we have $\nm{\othetnl-\othetl} \leq \log 2$.

\paragraph{Gathering the terms.} By gathering the results of the previous paragraphs
\[L(\othetnl) - L(\othet) ~~~\leq~~~ \psi(\tla + \log 2)~~(~1/\dikin(\tla)~\biass_\la ~~+~~ e^{\tla /2}/\dikin(\log 2)~\bvar_\la~)^2\]
Using the fact that $(a+b)^2 \leq 2a^2 + 2 b^2$, we have the desired result, with 
\eqal{\label{eq:Kbias-Kvar}
K_{\textup{bias}}(\tla) = 2\psi(\tla + \log 2)/\dikin(\tla)^2  ,~~K_{\textup{var}}(\tla) = 2 \psi(\tla + \log 2)e^{\tla}/\dikin(\log 2)^2.
}
which are bounded in \cref{df:constantsmain} and \cref{prp:constantsmain} of \cref{sec:app-main-result}.
\epr

\subsection{Analytic bounds for terms related to the variance}\label{sec:main-analytic-related}

In this lemma, we aim to control the essential supremum and the variance of the random vector $\Hl^{-1/2}(\othetl)\nabla \ell^\la_z(\othetl)$  relating it to quantities at $\othet$. The results will be used to control the variance via  Bernstein  concentration inequalities, so we are going to control its essential supremum and its variance.

\blm[Control of $\Hl(\othetl)^{-1/2}\nabla \ell^\la_z(\othetl)$]\label[lemma]{lm:bound-eff-dim} 
For any $0 < \la \leq \Bts$, we have 
\begin{enumerate}
    \item A bound on the essential supremum:
   
\[\sup_{z \in \supp(\rho)}\|\nabla \ell_z(\othetl)\|_{\Hl^{-1}(\othetl)} \leq \square_1~ \frac{\Bos}{\sqrt{\la}} + 2\square_2~\frac{\Bts }{\la} \biass_\la.\]

\item A bound on the variance
$$\Exp{\|\nabla \ell^\la_z(\othetl)\|^2_{\Hl^{-1}(\othetl)}}^{1/2} \leq \square_1 ~\sqrt{\Deff_\la} + \sqrt{2} \square_2~ \sqrt{\frac{\Bts }{\la}} \biass_\la,$$

\end{enumerate}

where $\square_1,\square_2$ are increasing functions of $\tla$ :
$\square_1(\tla) = e^{\tla/2}$
$\square_2(\tla) = e^{\tla/2} \left(1 + e^{\tla}\right)$.
\elm

\bpr
Start by noting that if  $\la \leq \Bts $, then $\sup_{z \in \supp(\rho)} \|\Hl^{-1/2}(\othet) \nabla^2 \ell^\la_z(\othet)^{1/2}\|^2 \leq 1 + \frac{\Bts}{\la} \leq 2 \frac{\Bts}{\la} $. 
Moreover, note that for any vector $h \in \hh$, multiplying and dividing by $\nabla^{2}\ell_z(\othet)^{1/2}$, \eqals{\|h\|_{\Hl^{-1}(\othet)} &:= \|\Hl^{-1/2}(\othet)~h\| = \|\Hl^{-1/2}(\othet) ~\nabla^{2}\ell_z(\othet)^{1/2}~\nabla^{2}\ell_z(\othet)^{-1/2}~h\| \\
    &\leq \|\Hl^{-1/2}(\othet) ~\nabla^{2}\ell_z(\othet)^{1/2}\|~\|\nabla^{2}\ell_z(\othet)^{-1/2}~h\| \\
    & \leq \sqrt{\frac{2\Bts}{\la}} \|\nabla^{2}\ell_z(\othet)^{-1/2}~h\|\\
    &=  \sqrt{\frac{2\Bts}{\la}} \|h\|_{\nabla^{2}\ell_z(\othet)^{-1}} \ \ ,}
    where the last bound is mentioned at the beginning of the proof. 
    Similarly, we can show
    \eqal{\label{eq:controldif}
    \|h\|_{\nabla^2 \ell_z(\othet)} \leq \sqrt{\frac{2\Bts}{\la}}~\|h\|_{\Hl(\othet)},\qquad
    \|h\|_{\Hl^{-1}(\othet)} \leq \sqrt{\frac{2\Bts}{\la}} \|h\|_{\nabla^{2}\ell_z(\othet)^{-1}} \ \ .
    }

\noindent{\normalfont \bfseries Essential supremum.}  Let $z \in \supp(\rho)$.  
    First note that using \cref{hess_control}, we have 
    \[\|\nabla \ell_z^{\la}(\othetl)\|_{\Hl^{-1}(\othetl)}  \leq  e^{\tla/2}\|\nabla \ell_z^{\la}(\othetl)\|_{\Hl^{-1}(\othet)}.\]
    Now bound
    \[\|\nabla \ell_z^{\la}(\othetl)\|_{\Hl^{-1}(\othet)} \leq  \|\nabla \ell_z^{\la}(\othetl) - \nabla \ell^\la_z(\othet)\|_{\Hl^{-1}(\othet)} + \|\nabla \ell_z^{\la}(\othet)\|_{\Hl^{-1}(\othet)}.\]
    Since $\nabla \ell_z^{\la}(\othet) = \nabla \ell_z(\othet) + \la \othet$, the last term is bounded by 
    $$\biass_\la + \sup_{z \in \supp(\rho)}\|\nabla \ell_z(\othet)\|_{\Hl^{-1}(\othet)} \leq \biass_\la + \frac{\Bos}{\sqrt{\la}}.$$
    For the first term, start by using \cref{eq:controldif}.
    \[\|\nabla \ell_z^{\la}(\othetl) - \nabla \ell^\la_z(\othet)\|_{\Hl^{-1}(\othet)} \leq \sqrt{\frac{2 \Bts}{\la}}~\|\nabla \ell_z^{\la}(\othetl) - \nabla \ell^\la_z(\othet)\|_{\nabla^2 \ell_z^\la(\othet)^{-1}}.\]
    
    Using \cref{grad_control2} on $\ell_z^\la$, we find
    \[\|\nabla \ell_z^{\la}(\othetl) - \nabla \ell^\la_z(\othet)\|_{\nabla^2 \ell_z^\la(\othet)^{-1}} \leq \dikins(\tla)~\|\othetl - \othet\|_{\nabla^2 \ell_z^\la(\othet)}.\]
    Applying once again \cref{eq:controldif}, we bound
    \[\|\othetl - \othet\|_{\nabla^2 \ell_z^\la(\othet)} \leq \sqrt{\frac{2\Bts}{\la}} ~\|\othetl - \othet\|_{\Hl(\othet)}.\]
    Finally, using \cref{grad_control1} on $\Ll$, we get
    \[\|\othetl - \othet\|_{\Hl(\othet)} \leq \frac{1}{\dikin(\tla)} \biass_\la.\]
    Hence, putting things together, we get
    \[\|\nabla \ell_z^{\la}(\othetl) - \nabla \ell^\la_z(\othet)\|_{\Hl^{-1}(\othet)} \leq \frac{2 \Bts}{\la} \frac{\dikins(\tla)}{\dikin(\tla)} \biass_\la = \frac{2 \Bts}{\la} e^{\tla} \biass_\la.\]
    
    We the combine all our different computation to get the bound.

\noindent{\normalfont \bfseries Variance.} We start by using \cref{hess_control} to show that 
    \[\Exp{\|\nabla \ell^\la_z(\othetl)\|^2_{\Hl^{-1}(\othetl)}}^{1/2} \leq e^{\tla/2} \Exp{\|\nabla \ell^\la_z(\othetl)\|^2_{\Hl^{-1}(\othet)}}^{1/2} .\]
    Then we use the triangle inequality 
    \[\Exp{\|\nabla \ell_z^{\la}(\othetl)\|^2_{\Hl^{-1}(\othet)}}^{1/2} \leq  \Exp{\|\nabla \ell_z^{\la}(\othetl) - \nabla \ell^\la_z(\othet)\|^2_{\Hl^{-1}(\othet)}}^{1/2} + \Exp{ \|\nabla \ell_z^{\la}(\othet)\|^2_{\Hl^{-1}(\othet)}}^{1/2}.\]
    We can easily bound the last term on the right hand side by $\biass_\la + \Deff_\la$. For the first term, we proceed as in the previous case to obtain
    \[\forall z \in \supp(\rho),~\|\nabla \ell_z^{\la}(\othetl) - \nabla \ell^\la_z(\othet)\|_{\Hl^{-1}(\othet)} \leq \sqrt{\frac{2\Bts}{\la}} \dikins(\tla) \|\othetl - \othet\|_{\nabla^2 \ell^\la_z(\othet)} .\]
    Now taking the expectancy of this inequality squared,
    \eqals{
    \Exp{\|\nabla \ell_z^{\la}(\othetl) - \nabla \ell^\la_z(\othet)\|^2_{\Hl^{-1}(\othet)}}^{1/2} &\leq \sqrt{\frac{2\Bts}{\la}} \dikins(\tla) \Exp{\|\othetl - \othet\|^2_{\nabla^2 \ell^\la_z(\othet)}}^{1/2}\\ &=\sqrt{\frac{2\Bts}{\la}} \dikins(\tla) \|\othetl - \othet\|_{\Hl(\othet)},
    }
    where the last equality comes from $\Exp{\nabla^2\ell^\la_Z(\othet)} = \Hl(\othet)$. Now applying \cref{grad_control1} to $L_{\la}$, we obtain
    \[\|\othetl - \othet\|_{\Hl(\othet)} \leq \frac{1}{\dikin(\tla)}\biass_\la.\]
    Regrouping all these bounds, we obtain 
    \[\Exp{\|\nabla \ell_z^{\la}(\othetl) - \nabla \ell^\la_z(\othet)\|^2_{\Hl^{-1}(\othet)}}^{1/2} \leq e^{\tla} \sqrt{\frac{2\Bts}{\la}}\biass_\la.\]
    Hence the final bound is proved, regrouping all our computations.

\epr

\subsection{Concentration lemmas}\label{sec:concentration_general}
Here we concentrate in high probability the quantities obtained in the analytical decomposition. Details on the proof technique are given in \cref{sec:proof} of the paper.

\blm[Equivalence of empirical and expected Hessian]\label[lemma]{lm:hessian_concentration}
Let $\theta \in \hh$ and $n \in \N$. For any $\delta \in (0,1]$, $\la > 0$, if 
\eqal{\label{eq:hessian_concbis}
n \geq 24 \frac{\Btwo(\theta)}{\la} \log \frac{8\Btwo(\theta)}{\la \delta } ,}
then with probability at least $1-\delta$: $\Hl(\theta) \preceq 2 \Hnl(\theta)$, or equivalently 
$$\|\Hl^{1/2}(\theta)\Hnl^{-1/2}(\theta)\|^2 \leq 2.$$
\elm
\bpr
By \cref{rem:apply-op-ineq} and the definition of $\Btwo(\theta)$, the condition we require on $n$ is sufficient to apply \cref{op_conc_2}, in particular \cref{eq:op_conc-Qhat_inv-Q}, to $\Hl(\theta), \Hnl(\theta)$, for $t=1/2$, which provides the desired result.
\epr

\blm[Concentration of the empirical gradient] \label[lemma]{lm:vector_concentration}
Let $n \in \N, \delta \in (0,1]$, $0 < \la \leq \Bts$. For any $k \geq 4$, if 
$n \geq k^2 \square_2^2 ~\frac{\Bts}{\la} \log \frac{2}{\delta}$, then with probability at least $1-\delta$, we have 
\eqal{\label{eq:emp-grad-conc-res}
 \|\nabla \Lnl(\othetl)\|_{\Hl^{-1}(\othetl)} \leq \frac{2\sqrt{3} }{k}\biass_\la + 2 \square_1~\sqrt{ \frac{\Deff_\la \vee (\Qs)^2~\log \frac{2}{\delta}}{n}}.
 }
Here, $\square_1,\square_2$ are defined in \cref{lm:bound-eff-dim} in \cref{sec:app-main-result} and $(\Qs)^2 = (\Bos)^2/\Bts$.
\elm
\bpr
\noindent{\normalfont \bfseries 1)} First let us concentrate $\Hl(\othetl)^{-1/2}\nabla \Lnl(\othetl)$ using a Bernstein-type inequality. \\\
    
    We can see $\Hl(\othetl)^{-1/2}\nabla \Lnl(\othetl)$ as the mean of $n$ i.i.d. random variables distributed from the law of the vector $\Hl(\othetl)^{-1/2}\nabla \ell_z(\othetl)$.
    
    As we have shown in \cref{lm:bound-eff-dim}, the essential supremum and variance of this vector is bounded, then we can use Bernstein inequality for random vectors \citep[e.g. Thm.~3.3.4 of][]{yurinsky1995sums}: for any $\la > 0$, any $n \in \N$ and $\delta \in (0,1]$, with probability at least $1 - \delta$, we have 
    $$ \|\Hl(\othetl)^{-1/2}\nabla \Lnl(\othetl)\| \leq \frac{2 M \log \frac{2}{\delta}}{n} + \sigma~\sqrt{\frac{2 \log \frac{2}{\delta}}{n}},$$
    where $M = \sup_{z \in \supp(\rho)} \|\nabla \ell_z(\othetl)\|_{\Hl^{-1}(\othetl)}$  and $\sigma = \Exp{\|\nabla \ell_z(\othetl)\|_{\Hl^{-1}(\othetl)}^2}^{1/2}$.\\\
    
\noindent{\normalfont \bfseries 2)} Using the bounds obtained in  \cref{lm:bound-eff-dim}, 
 
 \[M \leq \square_1~ \frac{\Bos}{\sqrt{\la}} + 2 \square_2~\frac{\Bts}{\la} \biass_\la,~~~~~\sigma \leq \square_1 ~\sqrt{\Deff_\la} + \sqrt{2}\square_2~ \frac{\sqrt{\Bts}}{\sqrt{\la}} \biass_\la.\]

\noindent{\normalfont \bfseries 3) } Injecting these in the Bernstein inequality, 

\eqals{ \|\Hl(\othetl)^{-1/2}\nabla \Lnl(\othetl)\| &\leq \frac{2 \left(\square_1~\Bos /\sqrt{\la}  + 2\square_2~ (\Bts/\la)\biass_\la\right)~ \log \frac{2}{\delta}}{n} \\
&+ \left(\square_1~ \Deff_\la^{1/2} + \sqrt{2} \square_2~\sqrt{\Bts/\la} \biass_\la\right)\sqrt{\frac{2 \log \frac{2}{\delta}}{n}} \\
& = \left[\frac{4 \square_2 ~\Bts~\log \frac{2}{\delta}}{\la n} + \sqrt{\frac{4\square_2^2 ~\Bts ~\log \frac{2}{\delta}}{\la n}}\right]\biass_\la\\
&+ \sqrt{\frac{2 \square_1^2~ \Deff_\lambda~\log \frac{2}{\delta}}{n}} + \sqrt{\frac{2~ \Bts~\log \frac{2}{\delta}}{\la n}} \sqrt{\frac{2 \square_1^2~(\Bos)^2/\Bts ~\log \frac{2}{\delta}}{n}}.
}
In the last inequality, we have regrouped the terms with a factor $\biass_\la$ and we have separated the first term of the decomposition in the following way : 
$$ \frac{2 \square_1 \Bos \log \frac{2}{\delta} }{\sqrt{\la} n} = \sqrt{\frac{2~ \Bts~\log \frac{2}{\delta}}{\la n}} \sqrt{\frac{2 \square_1^2~(\Qs)^2 ~\log \frac{2}{\delta}}{n}}.$$
Hence, we can bound the second line of the last inequality:
\[\sqrt{\frac{2 \square_1^2 \Deff_\lambda \log \frac{2}{\delta}}{n}} + \sqrt{\frac{2 \Bts \log \frac{2}{\delta}}{\la n}} \sqrt{\frac{2 \square_1^2 (\Qs)^2 \log \frac{2}{\delta}}{n}} \leq \left(1 + \sqrt{\frac{2  \Bts \log \frac{2}{\delta}}{\la n}} \right)\sqrt{\frac{2 \square_1^2  \Deff_\lambda \vee (\Qs)^2 \log \frac{2}{\delta}}{n}}.\]

Thus, if we assume that $n \geq k^2 \square_2^2 \frac{\Bts}{\la}\log \frac{2}{\delta}$, 
$$ \|\Hl(\othetl)^{-1/2}\nabla \Lnl(\othetl)\| \leq \left(\frac{4}{k^2}+\frac{2}{k}\right)\biass_\la + \left(1 + \frac{\sqrt{2}}{k}\right)\sqrt{\frac{2\square_1^2~\Deff_\la \vee (\Bos)^2/\Bts~\log \frac{2}{\delta}}{n}}.$$

In particular, for $k \geq 4$,
$$ \|\Hl(\othetl)^{-1/2}\nabla \Lnl(\othetl)\| \leq \frac{3}{k}\biass_\la + 2 \square_1 ~\sqrt{ \frac{\Deff_\la \vee (\Qs)^2~\log \frac{2}{\delta}}{n}}.$$

\epr

\blm [control of $\bvar_\la$]\label[lemma]{lm:bvar_control}
 Let $n \in \N$, $\delta \in (0,1]$ and $0 < \la \leq \Bts$. Assume that for a certain $k \geq 5$,
\[n \geq k^2 \square_2^2 \frac{\Bts}{\la}\log \frac{8 \square_1^2 \Bts}{\la \delta}.\]
Then with probability at least $1-2\delta$, we have 
$$\bvar_\la \leq \frac{6}{k}\biass_\la + 4 \square_1~\sqrt{ \frac{\Deff_\la \vee (\Qs)^2~\log \frac{2}{\delta}}{n}}.$$
Here, $\square_1,\square_2$ are defined in \cref{lm:bound-eff-dim}
\elm
\bpr
\begin{itemize}
    \item First we apply \cref{lm:hessian_concentration} to $\theta= \othetl$. Since $\Btwo(\othetl) \leq  e^{\tla}~\Bts = \square_1^2~\Bts$, we see that the condition 
\[n \geq 24 \frac{\Btwo(\othetl)}{\la} \log \frac{8 \Btwo(\othetl)}{\la \delta}\]
is satisfied if 
\[n \geq 24 \square_1^2 \frac{\Bts}{\la} \log \frac{8 \square_1^2 \Bts}{\la \delta}.\]
Because $k \geq 5$ and $\square_2 \geq \square_1$, and  we see that the assumption of this lemma imply the conditions above and hence \cref{lm:hessian_concentration} is satisfied. In particular, $\|\Hl(\othetl)^{1/2} \Hnl(\othetl)^{-1/2}\|^2 \leq 2$.
\item Note that the condition of this proposition also imply the conditions of \cref{lm:vector_concentration}, because $\la \leq \Bts$ and $\square_1 \geq 1$ imply $\frac{\square_1^2 \Bts}{\la \delta} \geq \frac{1}{\delta}$.
\end{itemize}
\epr

\subsection{Final results}\label{sec:final_results_general}

First, we find conditions on $n$ such that the hypothesis $\bvar_\la \leq \frac{\rla(\othetl)}{2}$ is satisfied. 

\blm \label[lemma]{lm:condition_gen_bound}
Let $n \in \N$, $\delta \in (0,1]$, $0 < \la \leq \Bts$ and
\[n \geq \triangle_1 \frac{\Bts}{\la}\log \frac{8 \square_1^2\Bts}{\la \delta},~~~n \geq \triangle_2~  \frac{\Deff_\lambda \vee (\Qs)^2}{\rla(\othet)^2} ~\log \frac{2}{\delta},\]
then with probability at least $1-2 \delta$
\[\frac{\bvar_\la}{\rla(\othetl)} \leq \square_1 \frac{\bvar_\la}{\rla(\othet)} \leq \frac{1}{2},\]
where $\square_1$, $\triangle_1,\triangle_2$ are constants defined in \cref{df:constantsmain}.
\elm

\bpr
Recall that $\tlab = \frac{\biass_\la}{\rla(\othet)}$. 

Using \cref{lm:bvar_control}, we see that under the conditions of this lemma, we have
\[ \square_1 \frac{\bvar_\la}{\rla(\othet)} \leq \frac{6 \square_1  ~\biass_\la}{k~\rla(\othet)} + 4 \square_1^2  \sqrt{ \frac{\Deff_\la \vee(\Qs)^2~\log \frac{2}{\delta}}{n~\rla(\othet)^2}}.\]\
Thus, taking $k = 24 \square_1 (1/2 \vee \tlab)$ and $n \geq 256 \square_1^4  \frac{\Deff_\la \vee \Qs}{\rla(\othet)^2} ~\log \frac{2}{\delta}$, both terms in the sum are bounded by $1/4$ hence the result.\\\

Note that here, we have defined 
\[\triangle_1 = 576 \square_1^2 \square_2^2 (1/2 \vee \tlab)^2,~~ \triangle_2 = 256 \square_1^4, \]
hence the constants in the definition above.
\epr

\bpr {\normalfont \bfseries of \cref{thm:general-result}}
First we recall that $\triangle_1$, $\triangle_2$, $\square_1$, $\mathsf{C}_{\textup{bias}}$ and $\mathsf{C}_{\textup{var}}$ are defined in \cref{df:constantsmain}, and bounded in \cref{prp:constantsmain}.

First note that, given the requirements on $n$, by \cref{lm:condition_gen_bound}, we have $\bvar_\la \leq \frac{\rla(\othetl)}{2}$ with probability at least $1-2\delta$. Thus, we are in a position to apply \cref{prp:anal_dec} : 
\eqals{
L(\othetnl) - L(\othet) \leq  K_{\textup{bias}}~ \biass_\la^2 + K_{\textup{var}}~\bvar_\la^2,
}
with $K_{\textup{bias}}, K_{\textup{var}}$ defined in the proof of the theorem.
Note that in the proof of \cref{lm:condition_gen_bound}, we have taken $k = 24 \square_1 (1/2 \vee \tlab) \geq 12$. Hence, using \cref{lm:bvar_control}, we find 
\[\bvar_\la \leq \frac{1}{2}\biass_\la + 4 \square_1~\sqrt{ \frac{\Deff_\la \vee (\Qs)^2~\log \frac{2}{\delta}}{n}}.\]

Hence, 
\[\bvar_\la^2 \leq \frac{1}{2}\biass_\la^2 + 32 \square_1^2 \frac{\Deff_\la \vee (\Qs)^2~\log \frac{2}{\delta}}{n}, \]
which yields the wanted result with $\mathsf{C}_{\textup{bias}} = K_{\textup{bias}} + \frac{1}{2}K_{\textup{var}}$ and $\mathsf{C}_{\textup{var}} = 32 \square_1^2 K_{\textup{var}}$.

\epr

\bpr  {\normalfont \bfseries of \cref{thm:general-result-weak}}

We get this theorem as a corollary of \cref{thm:general-result}. Indeed, $\forall \la \leq \Bts,~\Deff_\la \vee (\Qs)^* \leq \frac{(\Bos)^2}{\la}$, hence the result. 

\epr

\section{Explicit bounds for the simplified case}\label{sec:explicit-simple}

In this section, assume that \cref{asm:gen_scb,asm:bounded,asm:wd,asm:iid,asm:minimizer} hold.

Define the following constant $N$ : 

\eqal{\label{eq:constantNslow}  N = 36 A^2 \log^2\left(6 A^2\frac{1}{\delta}\right) ~~~ \vee ~~~ 256 \frac{1}{A^2} ~ \log \frac{2}{\delta} ~~~ \vee ~~~512 \left( \|\othet\|^2 \Rad^2\vee 1\right) \log \frac{2}{\delta} ,}
where $A = \frac{\Btb}{ \Bob }$.

We have the following slow rates theorem. 

\bt[Quantitative slow rates result]\label[theorem]{thm:quantitative_slow_rates}
Let $n \in \N$. Let $\delta \in (0,1]$.
Setting 
$$ \la = 16 
((\Rad\vee 1) \Bob)~ \frac{1}{\sqrt{n}} ~\log^{1/2} \frac{2}{\delta},$$
if $n \geq N$, with probability at least $1-2 \delta$,
\eqal{
L(\othetnl) - L(\othet) \leq
 48 ~~
\max(\Rad,1) \max(\|\othet\|^2,1) \Bob \frac{1}{\sqrt{n}} ~\log^{1/2} \frac{2}{\delta},
 }
and  $N = O\left(\poly(\Bob,\Btb,\Rad \|\othet\|)\right)$ is given explicitly in \cref{eq:constantNslow}. Here, $\poly$ denotes a certain rational function of the inputs.

\et

\bpr
Note that $\Deffsup_\la \leq \frac{\Bob^2}{\la}$. Hence, if $\la \leq \Btb$, then $\Deffsup_\la \vee (\Bob^2 / \Btb)\leq \frac{\Bob^2}{\la}$. 

\noindent{\normalfont \bfseries 1)}
Let us reformulate \cref{thm:gen_bound_22}.
Let $n \in \N$ and $0 < \la \leq \Btb $. Let $\delta \in (0,1]$. If 

\[ n \geq 512 \left( \|\othet\|^2 \Rad^2\vee 1\right) \log \frac{2}{\delta},~~~ n\geq 24 \frac{\Btb}{\la} \log \frac{8 \Btb}{\la \delta},~~~n \geq 16\triangle^2~\frac{\Rad^2\Bob^2}{ \la^2 } ~\log \frac{2}{\delta} , \]
then with probability at least $1 - 2 \delta$, 
\eqals{
L(\othetnl) - L(\othet) \leq \Cv~ \frac{\Bob^2}{\la  n}~\log \frac{2}{\delta} +\Cb \lambda \|\othet\|^2  ,}
where $\triangle,\Cb,\Cv$ are defined in \cref{rmk:constants2}.

\noindent{\normalfont \bfseries 2)}   Now setting $\la =16 \Rad \Bob  \log^{1/2}\frac{2}{\delta} \frac{1}{n^{1/2}}$, we see that the inequality
$$ n \geq 16\triangle^2~\frac{\Rad^2\Bob^2}{ \la^2 } ~\log \frac{2}{\delta} $$

is automatically satisfied since $\triangle \leq 4$.
Hence, if 
\[n \geq 512 \left( \|\othet\|^2 \Rad^2\vee 1\right) \log \frac{2}{\delta},~~~ n\geq 24 \frac{\Btb}{\la} \log \frac{8 \Btb}{\la \delta},~~~~0 < \la \leq \Btb,\]
then
\[L(\othetnl) - L(\othet) \leq \frac{\Cv}{256}~\frac{1}{\Rad^2} \la + \Cb \la \|\othet\|^2 \leq \left(\frac{\Cv}{256} + \Cb\right)\max(\frac{1}{\Rad^2},\|\othet\|^2) \la.\]

Since by \cref{rmk:constants2}, $\Cv \leq 84$ and $\Cb \leq 2$, we get
\[L(\othetnl) - L(\othet) \leq 3 \max(\frac{1}{\Rad^2},\|\othet\|^2) \la.\]

\noindent{\normalfont \bfseries 3)} Having our fixed $\la = 16  \frac{\Bob \Rad~\log^{1/2} \frac{2}{\delta}}{n^{1/2}}$, let us look for conditions for which
\[n \geq 512 \left( \|\othet\|^2 \Rad^2\vee 1\right) \log \frac{2}{\delta},~~~ n\geq 24 \frac{\Btb}{\la} \log \frac{8 \Btb}{\la \delta},~~~~0 < \la \leq \Btb,\]
are satisfied. 

To deal with $ n\geq 24 \frac{\Btb}{\la} \log \frac{8 \Btb}{\la \delta}$, bound 
$$\frac{\Btb}{\la} \leq \frac{1}{16}\frac{\Btb}{ \Rad \Bob~\log^{1/2} \frac{2}{\delta}}~n^{1/2}\leq \frac{1}{8}\frac{\Btb}{ \Rad \Bob}~n^{1/2},$$
where we have used the fact that $\log ^{1/2} \frac{2}{\delta} \geq \frac{1}{2}$. 
apply \cref{technicalities2} with $a_1 = 3, a_2 = 1, A = \frac{\Btb}{ \Rad\Bob }$ to get the following condition:

\[n \geq 4a_1^2 A^2\log^2\left(\frac{2a_1 a_2 A^2}{\delta}\right),\]
which we express as
\[n \geq 36 A^2 \log^2\left(6 A^2 \frac{1}{\delta}\right).\]
To deal with the bound $\la  < \Btb$, we need only apply the definition to obtain  
\[n \geq 256 \frac{\Rad^2\Bob^2}{\Btb^2} ~ \log \frac{2}{\delta}.\]
Thus, we can concentrate all these bounds as $n \geq N$ where 
$$  N = 36 A^2 \log^2\left(6 A^2\frac{1}{\delta}\right) ~~~ \vee ~~~ 256 \frac{1}{A^2} ~ \log \frac{2}{\delta} ~~~ \vee ~~~512 \left( \|\othet\|^2 \Rad^2\vee 1\right) \log \frac{2}{\delta} ,$$
where $A = \frac{\Btb}{ \Rad \Bob }$. 

\noindent{\normalfont \bfseries 4) } Since $\Rad$ is only an upper bound, we can replace $\Rad$ by $\Rad \vee 1$. In this case, we see that $A \leq \frac{\Btb}{\Bob}$ and 
$\max (\frac{1}{\Rad \vee 1},
(\Rad \vee 1) \|\othet\|^2) \leq (\Rad \vee 1)(\|\othet\|\vee 1)^2$ hence the final bounds. 

\epr

\section{Explicit bounds for the refined case}\label{sec:explicit-general}

In this part, we continue to assume \cref{asm:gen_scb,asm:bounded,asm:wd,asm:iid,asm:minimizer}. We present a classification of distributions $\rho$ and show that we can achieve better rates than the classical slow rates.

\bd[class of distributions]
Let $\alpha \in [1,+\infty]$ and $r \in [0,1/2]$. \\\
We denote with $ \clas_{\alpha,r}$ the set of probability distributions $\rho$ such that there exists $\Lc,\Qc \geq 0$, 
\begin{itemize}
    \item $\biass_\la \leq \Lc ~\lambda^{\frac{1 + 2r}{2}}$
    \item $\Deff_\la \leq \Qc^2~ \la^{-1/\alpha}$,
\end{itemize}
where this holds for any $0<\la \leq 1$. 
For simplicity, if $\alpha = +\infty$, we assume that $\Qc \geq \Qs$.
\ed

 Note that given our assumptions, we always have 

\eqal{\label{eq:basic_class}\rho \in \clas_{1,0},~~~  \Lc = \|\othet\|,~\Qc = \Bos.}

We also define 
\eqal{ \label{eq:lambda1}\la_1 = \left(\frac{\Qc}{\Qs}\right)^{2\alpha} \wedge 1 , }
such that 
$$ \forall \la \leq \la_1,~ \Deff_\la \vee (\Qs)^2 \leq \frac{\Qc^2}{\la^{1/\alpha}} .$$

\paragraph{Interpretation of the classes}

\begin{itemize}
    \item The bias term $\biass_\la$ characterizes the regularity of the objective $\othet$. In a sense, if $r$ is big, then this means $\othet$ is very regular and will be easier to estimate. The following results reformulates this intuition.

\br[source condition]\label[remark]{rmk:source_cond}
Assume there exists $0 \leq r \leq 1/2$ and $v \in \hh$ such that 
\[P_{\Ho(\othet)}\othet = \Ho(\othet)^r v.\]
Then we have 
\[\forall \la >0,~\biass_\la \leq \Lc~\la^{\frac{1 + 2r}{2}},~~~~\Lc = \|\Ho(\othet)^{-r}\othet\|.\]
\er

 \item The effective dimension $\Deff_\la$ characterizes the size of the space $\hh$ with respect to the problem. The higher $\alpha$, the smaller the space. If $\hh$ is finite dimensional for instance, $\alpha = +\infty$.
 
 \end{itemize}

 We will give explicit bounds for the performance of $\othetnl$ depending on which class $\rho$ belongs to, i.e., as a function of $\alpha,r$.

\paragraph{Well -behaved problems}

$\rla(\othet)$ has a limiting role. However, as soon as we have some sort of regularity, this role is no longer limiting, i.e. this quantity does not appear in the final rates and the constants in these rates have no dependence on the problem. This motivates the following definition.

We say that a problem is well behaved if the following equation holds.

\eqal{\label{eq:wb} \forall \delta \in (0,\frac{1}{2}],~ \exists \lambda_0(\delta) \in (0,1],~\forall 0< \lambda \leq \lambda_0(\delta),~\frac{\Lc \la^{1/2 + r}}{\rla(\othet)}~\log \frac{2}{\delta} \leq \frac{1}{2}.}

\br[well-behaved problems]\label[remark]{rmk:wb}
Note that \cref{eq:wb} is satisfied if one of the following holds.
\begin{itemize}
    \item If $\Rad =0$, then the condition holds for $\la_0 = 1$. 
    \item If $r > 0$, then the condition holds for $\la_0 = (2\Lc \Rad \log \frac{2}{\delta})^{-1/r} \wedge 1$.
    \item If there exists $\mu \in [0,1)$ and $\Fc \geq 0$ such that $\rla(\othet) \geq \frac{1}{\Fc}\la^{\mu/2}$, then this holds for $\la_0 = (2 \Rad \Fc \log \frac{2}{\delta})^{-2/(1-\mu + 2r)} \wedge 1$.
\end{itemize}
Moreover, if \cref{eq:wb} is satisfied, than for any $\la \leq \la_0$, $\tla \leq \log 2$.
\er

Note that the first possible condition corresponds to the case where the loss functions are quadratic in $\theta$ (if the loss is the square loss for instance). The second condition corresponds to having a strict source condition, i.e. something strictly better than just $\othet \in \hh$. Finally, the third condition corresponds to the fact that the radius $\rla$ decreases slower than the original bound of $\rla \geq \frac{\la^{1/2}}{\Rad}$, and hence it is not limiting.

Note that a priori, using only the assumptions, our problems do not satisfy \cref{eq:wb} (see \cref{eq:basic_class}, and the fact that $\rla \geq \frac{\sqrt{\la}}{\Rad}$).\\\

\subsection{Quantitative bounds}

In this section, for any given pair $(\alpha,r)$ characterizing the regularity and size of the problem, we associate 
\[\beta = \frac{1}{1 + 2r + 1/\alpha},~~~~\gamma = \frac{\alpha(1+2r)}{\alpha(1+2r) + 1}.\]

In what follows, we define 

\eqal{\label{eq:Nvalue}N = \frac{256 \Qc^2}{\Lc^2}\left(\Bts \wedge \la_0 \wedge \la_1\right)^{-1/\beta} ~~~\vee~~~\left(1296 \frac{1}{1-\beta}  A \log  \left(5184   \frac{1}{1-\beta}A^2 \frac{1}{\delta}\right)\right)^{1/(1-\beta)},}
where $A = \frac{\Bts \Lc^{2\beta} }{ \Qc^{2\beta} }$, $\la_0$ is given by \cref{eq:wb} and $\la_1$ is given by \cref{eq:lambda1} : $\la_1 = \frac{\Qc^{2\alpha}}{(\Qs)^{2\alpha}}$.

\bt[Quantitative results when \cref{eq:wb} is satisfied and $\alpha < \infty$ or $r > 0$]\label[theorem]{thm:eh1}
Let $\rho \in \clas_{\alpha,r}$  and that we have either $\alpha <\infty$ or $r>0$. Let $\delta \in (0,\frac{1}{2}]$.\\\
If \cref{eq:wb} is satisfied, and 
    \[n \geq N,~~~\lambda = \left(256~ \left(\frac{\Qc}{\Lc}\right)^{2}~\frac{1}{n}\right)^{\beta},\]
    then with probability at least $1-2 \delta$,
    \[\Lo(\othetnl) - \Lo(\othet) \leq 8~(256)^{\gamma}~\left(\Qc^{\gamma} ~ 
    \Lc^{1-\gamma}\right)^2 ~~\frac{1}{n^{\gamma}} \log \frac{2}{\delta},\]
    where $N$ is defined in \cref{eq:Nvalue}.

\et

\bpr 

Using the definition of $\la_1$, as soon as $\la \leq \la_1$ we have $\Deff_\la \vee (\Qs)^2 \leq \Qc^2 \la^{-1/\alpha}$.\\\

Let us formulate  \cref{thm:general-result} using the fact that $\rho \in \clas_{\alpha,r}$.\\\

Let $\delta \in (0,1]$, $0 < \la \leq \Bts \wedge \la_1$ and $n \in \N$ such that
\[n \geq \triangle_1 \frac{\Bts}{\la}\log \frac{8 \square_1^2\Bts}{\la \delta},~~~n \geq \triangle_2~  \frac{\Qc^2}{\la^{1/\alpha}\rla(\othet)^2} ~\log \frac{2}{\delta},\]
then with probability at least $1-2 \delta$
$$\L(\othetnl) - \L(\othet) \leq \Cb~\Lc^2\la^{1+2r} + \Cv~\frac{\Qc^2}{\la^{1/\alpha} n}~\log \frac{2}{\delta},
$$

where $\Cb,\Cv$ are defined in \cref{df:constantsmain}.
Now let us distinguish the two cases of our theorem.

\paragraph{Assume that $\rho$ satisfies \cref{eq:wb}}.
In this case the proof proceeds as follows. Note that as soon as $\la \leq \la_0$, we have $\frac{\biass_\la}{\rla(\othet)} \leq \frac{1}{2}$ and hence the bounds in \cref{prp:constantsmain} apply.

\noindent{\normalfont \bfseries 1)} First, we find a simple condition to guarantee 
$$\rla(\othet)^2\lambda^{ 1/\alpha} \geq \triangle_2~  \Qc^2
\frac{1}{n} ~\log \frac{2}{\delta}.$$

Using the fact that \cref{eq:wb} is satisfied, we see that if $\lambda \leq  \lambda_0$, then $\rla \geq 2\Lc \la^{1/2 + r} \log \frac{2}{\delta}$. Hence, this condition is satisfied if 
$$ \la \leq \la_0,~~~~4 \Lc^2 \la^{1+ 2r + 1/\alpha} \geq \triangle_2~  \Qc^2
\frac{1}{n} .$$ 

\noindent{\normalfont \bfseries 2) } Now fix $\Cla = 256 \geq \triangle_2/4$ (see \cref{prp:constantsmain}) and fix 
$$\lambda^{1 + 2r + 1/\alpha} =\Cla \frac{\Qc^2}{ \Lc^2} ~\frac{1}{n} \Longleftrightarrow \lambda = \left(\Cla \frac{\Qc^2}{ \Lc^2} ~\frac{1}{n}\right)^{\beta} .$$
where $\beta = 1/(1 + 2r + 1/\la) \in [1/2,1)$.

Using our restatement of \cref{thm:general-result},  
we have that with probability at least $1-2 \delta$,
$$\L(\othetnl) - \L(\othet) \leq 
  \left(\Cb + \frac{1}{\Cla}\Cv \log \frac{2}{\delta}\right)~\Lc^2 \lambda^{1 + 2r} \leq    \Kfinal~\log \frac{2}{\delta}~\Lc^2 \lambda^{1 + 2r},$$
  where we have set $\Kfinal = \left(\Cb + \frac{1}{256}\Cv\right) \leq 8$ (see \cref{prp:constantsmain}). \\\
  This result holds provided
     \eqal{\label{eq:cond_refor}0 < \lambda \leq \Bts \wedge \lambda_0 \wedge \la_1,~n \geq \triangle_1 \frac{\Bts}{\la}\log \frac{8 \square_1^2\Bts}{\la \delta}.}
     Indeed, we have shown in the previous point that since $\Cla \geq \frac{\triangle_2}{4}$, $\rla(\othet)^2\lambda^{ 1/\alpha} \geq \triangle_2~  \Qc^2
\frac{1}{n} ~\log \frac{2}{\delta}.$

\noindent{\normalfont \bfseries 3) }  Let us now work to guarantee the conditions in \cref{eq:cond_refor}.  \\\
First, to guarantee $n \geq \triangle_1 \frac{\Bts}{\la}\log \frac{8 \square_1^2\Bts}{\la \delta}$, bound
\[\frac{\Bts}{\la} = \frac{\Bts \Lc^{2\beta} n^{\beta}}{\Cla^{\beta} \Qc^{2\beta} ~\log^{\beta} \frac{2}{\delta}} \leq \frac{2}{\Cla^{\beta}}~\frac{\Bts \Lc^{2\beta} }{ \Qc^{2\beta} } n^{\beta}.\]

Then apply \cref{technicalities} with $a_1 = \frac{2\triangle_1}{\Cla^{\beta}}$, $a_2 = \frac{16 \square_1^2}{\Cla^{\beta}}$, $A = \frac{\Bts \Lc^{2\beta} }{ \Qc^{2\beta} }$. Since $\beta \geq 1/2$, using the bounds in \cref{prp:constantsmain}, we find $a_1 \leq 648$ and $a_2 \leq 4,$ hence the following sufficient condition:

\[ n \geq  \left(1296 \frac{1}{1-\beta}  A \log  \left(5184   \frac{1}{1-\beta}A^2 \frac{1}{\delta}\right)\right)^{1/(1-\beta)}.\]
  
 Then, to guarantee the condition 
\[ \lambda \leq \Bts \wedge \lambda_0 \wedge \la_1, \]
we simply need 
\[n \geq \frac{256 \Qc^2}{\Lc^2}\left(\Bts \wedge \la_0 \wedge \la_1\right)^{-1/\beta}.\]
Hence, defining 
$$N = \frac{256 \Qc^2}{\Lc^2}\left(\Bts \wedge \la_0 \wedge \la_1\right)^{-1/\beta} ~~~\vee~~~\left(1296 \frac{1}{1-\beta}  A \log  \left(5184   \frac{1}{1-\beta}A^2\frac{1}{\delta}\right)\right)^{1/(1-\beta)},$$
where $A = \frac{\Bts \Lc^{2\beta} }{ \Qc^{2\beta} }$, we see that as soon as $n \geq N$, \cref{eq:cond_refor} holds.

\epr

We now state the following corollary, for $r >0$. We define $N$ in the following way:
\eqal{\label{defNr} N = \frac{256 \Qc^2}{\Lc^2}\left(\Bts \wedge \la_0 \wedge \la_1\right)^{-1/\beta} ~~~\vee~~~\left(1296 \frac{1}{1-\beta}  A \log  \left(5184   \frac{1}{1-\beta}A^2\frac{1}{\delta}\right)\right)^{1/(1-\beta)}}
where $A = \frac{\Bts \Lc^{2\beta} }{ \Qc^{2\beta} }$, $\la_0 = (2 \Lc \Rad \log \frac{2}{\delta})^{-1/r}\wedge 1  $  and $\la_1 = \frac{\Qc^{2\alpha}}{(\Qs)^{2\alpha}}$.

\bcor\label[corollary]{cor:bounds_source_capacity_aux}
Assume $\rho \in \clas_{\alpha,r}$ with $r>0$. Let $\delta \in (0,0.5]$ and $n \geq N$, where $N$ is defined in \cref{defNr}. For
\[\lambda = \left(256~ \left(\frac{\Qc}{\Lc}\right)^{2}~\frac{1}{n}\right)^{\beta},\]
     with probability at least $1-2 \delta$,
    \[\Lo(\othetnl) - \Lo(\othet) \leq 8~(256)^{\gamma}~\left(\Qc^{\gamma} ~ 
    \Lc^{1-\gamma}\right)^2 ~~\frac{1}{n^{\gamma}} \log \frac{2}{\delta},\]
Moreover,
$N = O\left(\poly\left(\Bos,\Bts,\Lc,\Qc,\Rad,\log \frac{1}{\delta}\right)\right)$, which means that $N$ is bounded by a rational function of the arguments of $\poly$. 
\ecor

\bpr {\normalfont \bfseries of \cref{cor:bounds_source}}
We simply apply \cref{cor:bounds_source_capacity_aux} for $\alpha = 1$ and $\Qc = \Bos$.
\epr

\section{Additional lemmas}\label{sec:additional-lemmas}

\subsection{Self-concordance, sufficient conditions to define $L$ and related quantities }\label{sec:lemmas-self}

In this section, we will consider an arbitrary probability measure $\mu$ on $\Z$. We assume that $\ell_z$ satisfies \cref{asm:gen_scb} with a certain given function $\varphi$. Recall that $\Rad^{\mu} = \sup_{z \in \supp(\mu)} \sup_{g \in \varphi(z)} \|g\|$. In this section, we will also assume that $\Rad^{\mu} < \infty$.

\blm[Gronwall lemma]\label[lemma]{lm:gronwall}
Let $\varphi : \R \rightarrow \R$ be a differentiable function such that 
\[\forall t \in \R,~ \varphi^{\prime}(t) \leq C \varphi(t).\]
Then 
\[\forall (t_0,t_1)  \in \R^2,~\varphi(t_1) \leq e^{C|t_1 - t_0|} \varphi(t_0).\]
\elm

\blm\label[lemma]{lm:b2exp} 
Assume that there exists $\theta_0$ such that $\sup_{z \in \supp(\mu)}\Tr\left(\nabla^2\ell_z(\theta_0)\right) < \infty$
\begin{itemize}
\item $\sup_{z \in \supp(\mu)}\Tr\left(\nabla^2\ell_z(\theta)\right) < \infty$ for any $\theta \in \hh$;
\item For any given radius $T >0$, and any $\|\theta_0 \| \leq T$, we have 
\[\forall \|\theta\| \leq T, \forall z \in \Z,~ \Tr\left(\nabla^2\ell_z(\theta)\right) \leq \exp\left(2 \Rad^{\mu}T\right)\Tr\left(\nabla^2\ell_z(\theta_0)\right) < \infty. \]
\end{itemize}

\elm

\bpr

 Let $z \in \supp(\mu)$ be fixed. Using the same reasoning as in the proof of \cref{hess_control}, we can show 
$$\forall \theta_0,\theta_1 \in \hh,~\nabla^2 \ell_z(\theta_1) \preceq \exp\left(\sup_{g \in \varphi(z)}|g\cdot(\theta_1-\theta_0)|\right)\nabla^2 \ell_z(\theta_0) \preceq  \exp\left(\Rad^{\mu} \|\theta_1 - \theta_0\|\right)\nabla^2 \ell_z(\theta_0)$$
Where we have used the fact that $\Rad^{\mu} = \sup_{z \in \supp(\mu)}\sup_{g \in \varphi(z)}\|g\| < \infty$
Thus, in particular
$$\forall z \in \supp(\mu),~\forall \theta_0,\theta_1 \in \hh,~\Tr\left(\nabla^2 \ell_z(\theta_1)\right) \leq   \exp\left(\Rad^{\mu} \|\theta_1 - \theta_0\|\right)\Tr\left(\nabla^2 \ell_z(\theta_0)\right),$$
which leads to the desired bounds.

\epr

\blm\label[lemma]{lm:b1exp}
Assume that there exists $\theta_0$ such that 
$$\sup_{z \in \supp(\mu)}\Tr\left(\nabla^2\ell_z(\theta_0)\right) < \infty, \quad \sup_{z \in \supp(\mu)}\|\nabla \ell_z(\theta_0)\| < \infty.$$
Then
\begin{itemize}
\item $\sup_{z \in \supp(\mu)}\|\nabla \ell_z(\theta)\| < \infty$ for any $\theta \in \hh$
\item For any $T > 0$ and any $\|\theta_0\|, \| \theta\|\leq T, z \in 
\supp(\mu)$, 
\eqals{\|\nabla \ell_z(\theta)\| &\leq \|\nabla \ell_z(\theta_0)\|+ 2T \Tr\left(\nabla^2\ell_z(\theta_0)\right) \\
&  + 4 \Rad^{\mu}~\psi( 2 \Rad^{\mu}T)~\Tr\left(\nabla^2\ell_z(\theta_0)\right)~R^2.}
\end{itemize}

\elm

\bpr

Fix $z \in \Z$, $\theta_0,\theta_1 \in \hh$ and $h \in \hh$. 
Let us look at the function 
$$f : t \in [0,1] \mapsto \left(\nabla \ell_z(\theta_t) - \nabla \ell_z(\theta_0) - t\nabla^2 \ell_z(\theta_0)(\theta_1 - \theta_0)\right) \cdot  h.$$
We have $f^{\prime \prime}(t) = \nabla^3 \ell_z(\theta_t)[\theta_1 - \theta_0,\theta_1-\theta_0,h]$. 
By the self-concordant assumption, we have 
\eqals{\left|f^{\prime \prime}(t)\right| &\leq \sup_{g \in \varphi(z)}|g\cdot h| \nabla^2 \ell_z(\theta_t)[\theta_1 - \theta_0,\theta_1 - \theta_0] \\
&\leq \sup_{g \in \varphi(z)}|g\cdot h|  \exp(t \sup_{g \in \varphi(z)}|g \cdot \theta_1 - \theta_0|) \|\theta_1 - \theta_0\|^2_{\nabla^2 \ell_z(\theta_0)}.}
Integrating this knowing $f^{\prime}(0) = f(0) = 0$ yields
$$|f(1)| \leq \sup_{g \in \varphi(z)} |g\cdot h|~\psi(\sup_{g \in \varphi(z)}|g\cdot (\theta_1 - \theta_0)|)\|\theta_1 - \theta_0\|^2_{\nabla^2 \ell_z(\theta_0)}.$$
Hence :
\[\|\nabla \ell_z(\theta_1) - \nabla \ell_z(\theta_0)\| \leq \|\nabla^2 \ell_z(\theta_0)\|~ \|\theta_1 - \theta_0\| + \|\varphi(z)\|~\psi(\sup_{g \in \varphi(z)}|g\cdot (\theta_1 - \theta_0)|)~\|\nabla^2 \ell_z(\theta_0)\|~\|\theta_1 - \theta_0\|^2 \]
where $\psi(t) = (e^t - t- 1)/t^2$. 
Then, noting that $\|\nabla^2 \ell_z(\theta)\|\leq \Tr(\nabla^2 \ell_z(\theta))$, we have proved our lemma. 

\epr

\blm\label[lemma]{lm:b0exp}
Assume that there exists $\theta_0$ such that 
$$\sup_{z \in \supp(\mu)}\Tr\left(\nabla^2\ell_z(\theta_0)\right) < \infty, \quad \sup_{z \in \supp(\mu)}\|\nabla \ell_z(\theta_0)\| < \infty, \quad \sup_{z \in \supp(\mu)}|\ell_z(\theta_0)| < \infty.$$
Then 
\begin{itemize}
\item For any $\theta \in \hh$, $\sup_{z \in \supp(\mu)}|\ell_z(\theta)| < \infty$ 
\item For any $\theta_0 \in \hh$, $T \geq \|\theta_0\|, \|\theta\|\leq T,~ z \in \supp(\mu)$, we have:
\eqals{|\ell_z(\theta)|\leq |\ell_z(\theta_0)| + 2\|\nabla \ell_z(\theta_0)\|T  +  \psi(2\Rad^{\mu}~T) ~\Tr(\nabla^2 \ell_z(\theta_0))~T^2.}
\end{itemize}

\elm

\bpr
 Proceeding as in the proof of \cref{function_values_control}, we get 
\[\forall z \in \Z,~ \forall \theta_0,\theta_1 \in \hh, ~0 \leq  \ell_z(\theta_1) - \ell_z(\theta_0) - \nabla \ell_z(\theta_0)(\theta_1 - \theta_0) \leq \psi(\sup_{g \in \varphi(z)}|g\cdot(\theta_1 - \theta_0)|) \|\theta_1 - \theta_0\|_{\nabla^2 \ell_z(\theta_0)}^2\]
where $\psi(t) = (e^t - t- 1)/t^2$. 

\epr

To conclude, we give the following result. 

\bp\label[proposition]{prp:wd}
Let $\la \geq 0$. If a probability measure $\mu$ and $\ell$ satisfy \cref{asm:gen_scb,asm:bounded,asm:wd}, the function $L_{\mu,\la}(\theta) := \Expb{\mu}{\ell_z(\theta)} + \la \|\theta\|^2$ and $\nabla L_{\mu,\la}(\theta), \nabla^2 L_{\mu,\theta}(\theta)$ are well-defined for any $\theta \in \hh$, and we can differentiate under the expectation. Moreover,
$$\forall \theta \in \hh,~\sup_{z \in \supp(\rho)}|\ell_z(\theta)|,\sup_{z \in \supp(\rho)}\|\nabla \ell_z(\theta)\|,~\sup_{z \in \supp(\rho)}\Tr\left(\nabla^2\ell_z(\theta)\right) < \infty .$$
\ep

\bpr
We combine the results given in \cref{lm:b2exp,lm:b1exp,lm:b0exp}.
\epr

\subsection{Bernstein inequalities for operators}\label{sec:lemmas-operators}

We start by proposing a slight modification of Proposition 6 in \citep{Generalization}. First we need to introduce the following quantitity and some notation for Hermitian operators.
We denote by $\preceq$ is the partial order between positive semidefinite Hermitian operators. Let $A, B$ be bounded Hermitian operators on $\hh$,  
$$A \preceq B ~~~\Longleftrightarrow~~~ v \cdot (Av) \leq v \cdot (B v), ~~\forall v \in \hh ~~~\Longleftrightarrow ~~~ B-A ~\textrm{is positive semidefinite}.$$

Let $q$ be a random positive semi-definite operator and let $\mathbf{Q} := \Exp{q}$, denote by ${\cal F}(\la)$ the function of $\la$ defined as 
$$ {\cal F}(\la) := \textrm{ess}\sup \Tr\left(\mathbf{Q}_\la^{-1/2} q \mathbf{Q}_\la^{-1/2}  \right),$$
where $\textrm{ess}\sup$ is the {\em essential support} of $q$. 

\br\label[remark]{rem:Finfty-la}
Note that if $\Tr(q) \leq c_0$, for a $c_0 > 0$ almost surely, then  ${\cal F}(\la) \leq c_0/\la$. Vice versa, if ${\cal F}(\la_0) < \infty$ for a given $\la_0 > 0$, then $\Tr(q) \leq (\|Q\| + \la_0) {\cal F}(\la_0)$ almost surely, moreover ${\cal F}(\la) < \frac{\|\mathbf{Q}\| + \la_0}{\|\mathbf{Q}\| + \la}{ \cal F}(\la_0)$ for any $\la > 0$.
\er

\bp[Prop.~6 of \citep{Generalization}] \label[proposition]{op_concentration}
Let $q_1,...,q_n$ be identically distributed random positive semi-definite operators on a separable Hilbert space $\hh$ such that the $q$ are trace class and $\mathbf{Q} = \Exp{q}$. Let $\mathbf{Q}_n = \frac{1}{n}\sum_{i=1}^n{q_i}$ and take $0<\la \leq \|\mathbf{Q}\|$ and assume ${\cal F}(\la) < \infty$. For any $\delta > 0$, the following holds with probability at least $1 - \delta$:
\[\|\mathbf{Q}_\la^{-1/2}\left(\mathbf{Q} - \mathbf{Q}_n\right)\mathbf{Q}_\la^{-1/2}\| \leq \frac{2 \beta(1 + \Finf(\la))}{3n} + \sqrt{\frac{2 \beta \Finf(\la)}{n}},~~~~\beta =\log \frac{8 \Finf(\la)}{\delta} \]
\ep

\bpr
Use Proposition 3 of \citep{Generalization} and proceed as in the proof of Proposition 6 of \citep{Generalization} except that we bound $\Tr(\mathbf{Q}_\la^{-1}\mathbf{Q}) \leq \Finf(\la)$ instead of bounding $\Tr(\mathbf{Q}_\la^{-1}\mathbf{Q}) \leq \frac{\Tr(\mathbf{Q})}{\la}$, we find this result.
\epr

Here we slightly extend the results of Prop.~8 and Prop.~6 of \citep{Generalization}, to extend the range of $\la$ for which the result on the partial order between operators holds, from $0 < \la <\|\mathbf{Q}\|$ to $\la > 0$.
\bp[Prop.~8 together with Prop.~6 of \citep{Generalization}] \label[proposition]{op_conc_2}
Let $q_1,...,q_n$ be identically distributed random positive semi-definite operators on a separable Hilbert space $\hh$ such that the $q$ are trace class and $\mathbf{Q} = \Exp{q}$. Let $\mathbf{Q}_n = \frac{1}{n}\sum_{i=1}^n{q_i}$. Let any $\delta \in (0,1]$, $t >0, 0 < \la \leq \|\mathbf{Q}\|$ and assume ${\cal F}(\la) < \infty$, when  
\eqal{\label{eq:cond-op_conc_3}
n \geq 8 \Finf(\la) \log \frac{8 \Finf(\la)}{\delta} \left(\frac{1}{4t^2} + \frac{1}{t} \right) 
}

then the following holds with probability at least $1 -  \delta$:
\eqal{\label{eq:op_conc_res_0}
\|\mathbf{Q}_\la^{-1/2}\left(\mathbf{Q} - \mathbf{Q}_n\right)\mathbf{Q}_\la^{-1/2}\| \leq t.
}

Moreover let $\la > 0, \delta \in (0,1]$ and \cref{eq:cond-op_conc_3} is satisfied for $t \leq 1/2$,
then the following holds with probability at least $1-\delta$,
\eqal{\label{eq:op_conc-Qhat_inv-Q}
\mathbf{Q}_\la \preceq 2 \mathbf{Q}_{n,\la}, ~~~ \Longleftrightarrow ~~~ \|\mathbf{Q}_{n,\la}^{-1/2}\mathbf{Q}_\la^{1/2}\|^{2} \leq 2. 
}
Finally, let $\la > 0, \delta \in (0,1]$, \cref{eq:cond-op_conc_3} is satisfied for $t \leq 1/2$ and $$n \geq 16 \frac{c_0^2}{\|\mathbf{Q}\|^2} \log \frac{2}{\delta},$$ \
with $c_0 = \textrm{ess}\sup \Tr(q)$, then the following holds with probability at least $1-\delta$,
\eqal{\label{eq:op_conc_Qhat-Qinv}
\mathbf{Q}_{n,\la} \preceq  \frac{3}{2} \mathbf{Q}_\la, ~~~ \Longleftrightarrow ~~~  \|\mathbf{Q}_{n,\la}^{1/2}\mathbf{Q}_\la^{-1/2}\|^{2} \leq 3/2.
}
\ep
\begin{proof}

\noindent{\normalfont \bfseries Point 1)}
Let $\delta \in (0,1]$ and $0<\lambda\leq \mathbf{Q}$. Using \cref{op_concentration}, we have that with probability at least $1-\delta$,
\[\|\mathbf{Q}_\la^{-1/2}\left(\mathbf{Q} - \mathbf{Q}_n\right)\mathbf{Q}_\la^{-1/2}\| \leq \frac{2 \beta(1 + \Finf(\la))}{3n} + \sqrt{\frac{2 \beta \Finf(\la)}{n}},~~~~\beta =\log \frac{8 \Finf(\la)}{\delta}.\]
Now note that if $\lambda \leq \|\mathbf{Q}\|$, we have 
$$ \frac{1}{2} \leq \frac{\|\mathbf{Q}\|}{\|\mathbf{Q}\| + \lambda} = \|\mathbf{Q}_\la^{-1}\mathbf{Q}\| \leq \Tr\left(\mathbf{Q}_\la^{-1}\mathbf{Q}\right) \leq \Finf(\la).$$
Thus we can bound $1 + \Finf(\la) \leq 3\Finf(\la)$, and we rewrite the previous bound 
\[\|\mathbf{Q}_\la^{-1/2}\left(\mathbf{Q} - \mathbf{Q}_n\right)\mathbf{Q}_\la^{-1/2}\| \leq \frac{2 \beta \Finf(\la)}{n} + \sqrt{\frac{2 \beta \Finf(\la)}{n}},~~~~\beta =\log \frac{8 \Finf(\la)}{\delta}.\]

\noindent{\normalfont \bfseries Point 2)} Now let $t > 0$, $\delta \in (0,1]$ and $0< \lambda \leq \|\mathbf{Q}\|$. If 
\[n \geq 8 \Finf(\la) \beta \left(\frac{1}{4t^2} + \frac{1}{t} \right) , \]
then 
\[\|\mathbf{Q}_\la^{-1/2}\left(\mathbf{Q} - \mathbf{Q}_n\right)\mathbf{Q}_\la^{-1/2}\|  \leq t .\]
Indeed, assume we want to find $n_0 > 0$ for which for all $n\geq n_0,~~\frac{A}{n} + \sqrt{\frac{B}{n}} \leq \frac{1}{2}$
where $A,B \geq 0$. setting $x = \sqrt{n}$, this is equivalent to finding $x_0$ such that $\forall x \geq x_0,~\frac{x^2}{2}- \sqrt{B}x - A \geq 0$. 
A sufficient condition for this is that $x \geq  \sqrt{B} + \sqrt{B + 2A}$. 
Thus, since $A,B \geq 0$, the condition $x \geq 2\sqrt{B+2A}$ is sufficient, hence the condition $n \geq 4(B + 2A)$. Then we apply this to the following $A$ and $B$ to obtain the condition.
\[A = \frac{\beta\Finf(\la)}{t},~B = \frac{\beta \Finf(\la)}{2t^2}.\]

\noindent{\normalfont \bfseries Point 3) } 
When $\la > \|\mathbf{Q}\|$, the result is obtained noting that
$$\|\mathbf{Q}_\la^{1/2}\mathbf{Q}_{n,\la}^{-1/2}\|^2 \leq \frac{\|\mathbf{Q}\| + \la}{\la} = 1 + \frac{\|\mathbf{Q}\|}{\la} \leq 2.$$
When, on the other hand $0 < \la \leq\|\mathbf{Q}\|$, the final result is obtained by applying Prop.~6 and Prop.~8 of \citep{Generalization}, or equivalently applying \cref{eq:op_conc_res_0}, with $t=1/2$, for which the following holds with probability $1-\delta$:  $\|\mathbf{Q}_\la^{-1/2}\left(\mathbf{Q} - \mathbf{Q}_n\right)\mathbf{Q}_\la^{-1/2}\| \leq t$ and noting that,
$$ \|\mathbf{Q}_\la^{1/2}\mathbf{Q}_{n,\la}^{-1/2}\|^2 \leq \frac{1}{1-\|\mathbf{Q}_\la^{-1/2}\left(\mathbf{Q} - \mathbf{Q}_n\right)\mathbf{Q}_\la^{-1/2}\|} \leq 2.$$
To conclude this point, we recall that, given two Hermitian operators $A, B$ and $t > 0$, the inequality $A \preceq t B$ is equivalent to $B^{-1/2} A B^{-1/2} \preceq t I$, when $B$ is invertible. Since $B^{-1/2} A B^{-1/2}$ and $t I$ are commutative, then $B^{-1/2} A B^{-1/2} \preceq t I$ is equivalent to $v \cdot (B^{-1/2} A B^{-1/2}v) \leq t \|v\|^2$ for any $v \in \hh$, which in turn is equivalent to $\|B^{-1/2} A B^{-1/2}\| \leq t$. So $$\|A^{1/2}B^{-1/2}\|^2 \leq t ~~\Longleftrightarrow~~ A \preceq t B.$$

\noindent{\normalfont \bfseries Point 4)}
First note that
\eqal{\label{eq:bound-op-left}
\|\mathbf{Q}_\la^{-1/2}\mathbf{Q}_{n,\la}^{1/2}\|^2 \leq 1 +  \|\mathbf{Q}_\la^{-1/2}\left(\mathbf{Q} - \mathbf{Q}_n\right)\mathbf{Q}_\la^{-1/2}\|.
}
When $0< \la \leq \|\mathbf{Q}\|$, by applying \cref{eq:op_conc_res_0} with $t=1/2$, we have with probability $1-\delta$: $\|\mathbf{Q}_\la^{-1/2}\left(\mathbf{Q} - \mathbf{Q}_n\right)\mathbf{Q}_\la^{-1/2}\| \leq t$, moreover
by \cref{eq:bound-op-left} we have
$$\|\mathbf{Q}_\la^{-1/2}\mathbf{Q}_{n,\la}^{1/2}\|^2 \leq 1 + t \leq 3/2.$$
When instead $\la > \|\mathbf{Q}\|$, we consider the following decomposition
\eqals{
\|\mathbf{Q}_\la^{-1/2}\left(\mathbf{Q} - \mathbf{Q}_n\right)\mathbf{Q}_\la^{-1/2}\| \leq \frac{1}{\la} \|\mathbf{Q} - \mathbf{Q}_n\| \leq \frac{1}{\la} \|\mathbf{Q} - \mathbf{Q}_n\|_{HS},
}
where we denote by $\|\cdot\|_{HS}$, the Hilbert-Schmidt norm (i.e. $\|A\|_{HS}^2 = \Tr(A^*A)$) and $\|\mathbf{Q} - \mathbf{Q}_n\|_{HS}$ is well defined since both $\mathbf{Q}, \mathbf{Q}_n$ are trace class. Now since the space of Hilbert-Schmidt operators on a separable Hilbert space is itself a separable Hilbert space and $q$ are bounded almost surely by $c_0 := \textrm{ess}\sup \Tr(q)$, we can concentrate $\|\mathbf{Q} - \mathbf{Q}_n\|_{HS}$ via Bernstein inequality for random vectors \citep[e.g. Thm.~3.3.4 of][]{yurinsky1995sums}, obtaining with probability at least $1-\delta$
$$ \|\mathbf{Q} - \mathbf{Q}_n\|_{HS} \leq \frac{2 c_0 \log \frac{2}{\delta}}{n} + \sqrt{\frac{2 c_0^2 \log \frac{2}{\delta}}{n}} \leq \|\mathbf{Q}\|/2,$$
where the last step is due to the fact that we require $n \geq 16 c_0^2(\log \frac{2}{\delta})/\|\mathbf{Q}\|^2$, and the fact that by construction $\|\mathbf{Q}\| \leq B$. Then, 
$$\|\mathbf{Q}_\la^{-1/2}\mathbf{Q}_{n,\la}^{1/2}\|^2 \leq 1 + \frac{\|\mathbf{Q}\|}{2\la} \leq {3/2}.$$
The final result on $\preceq$ is obtained as for Point 5.
\end{proof}  

\br\label[remark]{rem:apply-op-ineq}
Let $\Tr(q) \leq c_0$ almost surely, for a $c_0 > 0$. Then  ${\cal F}(\la) \leq c_0/\la$. So \cref{eq:cond-op_conc_3} is satisfied when
$$n \geq \frac{8 c_0}{\la} \log \frac{8 c_0}{\la~\delta} \left(\frac{1}{4t^2} + \frac{1}{t}\right),$$ 
since ${\cal F}(\la) \leq c_0/\la$ as observed in \cref{rem:Finfty-la}. In particular, when $t=1/2$, \cref{eq:cond-op_conc_3} is satisfied when
$$n \geq \frac{24 c_0}{\la} \log \frac{8 c_0}{\la~\delta} .$$ 
\er

\subsection{Sufficient conditions to bound $n$ in order to guarantee $n \geq C_1 n^{p} \log \frac{C_2 n^{p}}{\delta}$}

\blm \label[lemma]{technicalities2}
Let $a_1,a_2,A \geq 0$ and $\delta >0$. If
\[n \geq 4a_1^2 A^2\log^2\left(\frac{2 a_1 a_2 A^2}{\delta}\right),\]
then 
$n \geq a_1 A n^{1/2} \log \frac{a_2 A n^{1/2}}{\delta}$.
\elm

\bpr
Indeed, note that 
$$n \geq a_1 A n^{1/2} \log \frac{a_2 A n^{1/2}}{\delta} \Longleftrightarrow \frac{a_1 A }{n^{1/2}} \log \frac{a_2 A n^{1/2}}{\delta} \leq 1 .$$
Now use the fact that for $A,B \geq 0$, $k \geq 2 A \log (2AB)$ implies $\frac{A}{k}\log (Bk) \leq 1$.
Indeed, $\log(Bk) = \log(2AB) + \log \frac{Bk}{2AB} = \log(2AB) + \log \frac{k}{2A} \leq \log(2AB) + \frac{k}{2A}$.
Hence, multiplying by $\frac{A}{k}$, we get the result.

We apply this to $A = a_1 A, B = \frac{a_2 A}{\delta}$ and $k = n^{1/2}$ to get the bound.
\epr

\blm \label[lemma]{technicalities}
Let $a_1,a_2,A \geq 0$ and $\delta >0$. Let $p \in [\frac{1}{2},1)$. If 
\[n^{1-p} \geq 2 \frac{1}{1-p} a_1 A \log  \left(2 a_1 (a_2\vee 1)  \frac{1}{1-p}A^2 \frac{1}{\delta}\right),\]
then 
\[n \geq a_1 A n^p \log \frac{a_2 A n^p }{\delta}.\]
\elm

\begin{proof}
\noindent{\normalfont \bfseries 1)}
Let $C_1,C_2 \geq 0$, and $p \in [0,1)$. 
Then  
$$ n \geq C_1 n^p \log (C_2 n^p)\Longleftrightarrow \frac{C_1 \frac{p}{1-p}}{n^{1-p}} \log \left(C_2^{(1-p)/p} n^{1-p}\right) \leq 1.$$
Now use the fact that for $A,B \geq 0$, $k \geq 2 A \log (2AB)$ implies $\frac{A}{k}\log (Bk) \leq 1$ (see proof of \cref{technicalities2}).

Thus, $n^{1-p} \geq 2C_1 \frac{p}{1-p} \log  \left(2 C_1 \frac{p}{1-p}C_2^{(1-p)/p}\right)$ is a sufficient condition.

\noindent{\normalfont \bfseries 2)} Now taking $C_1 = a_1 A$ and $C_2 =\frac{a_2 A}{\delta}$, we find that 
\[n^{1-p} \geq 2 \frac{p}{1-p} a_1 A \log  \left(2 a_1 a_2^{(1-p)/p}  \frac{p}{1-p}A^{1/p} (\frac{1}{\delta})^{(1-p)/p}\right).\]
Since $0.5 \leq p \leq 1 $, we see that $\frac{1-p}{p} \leq 1$ and $\frac{1}{p} \leq 2$ and thus we get our final sufficient condition.
\[n^{1-p} \geq 2 \frac{1}{1-p} a_1 A \log  \left(2 a_1 (a_2\vee 1)  \frac{1}{1-p}A^2 \frac{1}{\delta}\right).\]

\end{proof}

\end{document}